\documentclass{article}
\usepackage{nips08submit_e,times}
\usepackage{epsf}
\usepackage{amsmath}
\usepackage{amssymb}
\usepackage{graphicx}
\usepackage{tabularx}
\usepackage{rotating}
\usepackage{algorithm}
\usepackage{algorithmic}
\usepackage{boxedminipage}
\usepackage{url}

\newtheorem{theorem}{Theorem}
\newtheorem{definition}[theorem]{Definition}
\newtheorem{proposition}[theorem]{Proposition}
\newtheorem{corollary}[theorem]{Corollary}
\newtheorem{lemma}[theorem]{Lemma}

\newtheorem{condition}[theorem]{Condition}
\newtheorem{remark}[theorem]{Remark}

\def\tps{^{\mathsf{T}}}
\DeclareMathOperator{\diag}{\ diag}
\DeclareMathOperator{\range}{range}
\DeclareMathOperator{\rank}{rank}

\DeclareMathOperator{\spann}{span}
\providecommand{\abs}[1]{\left|#1\right|}

\providecommand{\pnorm}[2]{\left\|#1\right\|_#2}
\providecommand{\infnorm}[1]{\left\|#1\right\|_\ensuremath{\infty}}
\providecommand{\Rbbvec}[1]{\ensuremath{\mathbb{R}^{#1}}}
\providecommand{\Rbbmat}[2]{\ensuremath{\mathbb{R}^{#1\times #2}}}

\def\btilde{{\widetilde{b}}}
\def\Btilde{{\widetilde{B}}}

\def\Utilde{{\widetilde{U}}}

\def\bhat{{\widehat{b}}}
\def\Bhat{{\widehat{B}}}
\def\Uhat{{\widehat{U}}}

\def\qhat{{\widehat{q}}}
\def\phat{{\widehat{p}}}

\def\bvec{{\vec{b}}}

\def\hvec{{\vec{h}}}

\def\lvec{{\vec{l}}}
\def\qvec{{\vec{q}}}
\def\pvec{{\vec{p}}}
\def\vvec{{\vec{v}}}
\def\xvec{{\vec{x}}}
\def\onevec{{\vec{1}}}
\def\onevect{{\vec{1}\tps}}

\def\cvec{{\vec{c}}}

\def\Bhatx{{\widehat{B}_x}}

\def\Btildex{{\widetilde{B}_x}}
\def\UO{{(U\tps O)}}

\def\UOR{{(U\tps OR)}}
\def\UORi{{(U\tps OR)^{-1}}}
\def\UhOR{{(\widehat{U}\tps OR)}}
\def\UhORi{{(\widehat{U}\tps OR)^{-1}}}
\def\Ptwoone{{P_{2,1}}}
\def\Pone{{\vec{P}_1}}
\def\Pthreexone{{P_{3,x,1}}}

\def\Pthreetwoone{{P_{3,2,1}}}
\def\Ph{{\widehat{P}}}
\def\Phtwoone{{\Ph_{2,1}}}
\def\Phone{{\Ph_1}}
\def\Phthreexone{{\Ph_{3,x,1}}}
\def\Phthreetwoone{{\Ph_{3,2,1}}}
\def\etwoone{{\epsilon_{2,1}}}
\def\eone{{\epsilon_{1}}}
\def\ethreexone{{\epsilon_{3,x,1}}}
\def\sPtwoone{{\sigma_k(P_{2,1})}}
\def\sOR{{\sigma_k(OR)}}
\def\shPtwoone{{\sigma_k(\widehat{P}_{2,1})}}

\def\delone{{\delta_{1}}}
\def\delinf{{\delta_{\infty}}}

\def\bone{{\bvec_1}}
\def\binf{{\bvec_\infty}}
\def\binft{{\bvec_\infty\tps}}

\def\bhone{{\bhat_1}}
\def\bhinf{{\bhat_\infty}}
\def\bhinft{{\bhat_\infty\tps}}

\def\btone{{\btilde_1}}
\def\btinf{{\btilde_\infty}}
\def\btinft{{\btilde_\infty\tps}}

\def\pivec{{\vec{\pi}}}
\def\nuvec{{\vec{\nu}}}

\def\phivec{{\vec{\phi}}}
\def\psivec{{\vec{\psi}}}
\def\xivec{{\vec{\xi}}}
\def\sigmavec{{\vec{\sigma}}}
\def\zetavec{{\vec{\zeta}}}

\numberwithin{equation}{section}
\title{Reduced-Rank Hidden Markov Models}

\author{
Sajid M.~Siddiqi\\
Robotics Institute\\
Carnegie Mellon University\\
Pittsburgh, PA 15213 \\
\texttt{siddiqi@cs.cmu.edu} \\
\And
Byron Boots \\
Computer Science Department \\
Carnegie Mellon University \\
Pittsburgh, PA 15213 \\
\texttt{beb@cs.cmu.edu} \\
\And
Geoffrey J.~Gordon \\
Machine Learning Department \\
Carnegie-Mellon University \\
Pittsburgh, PA 15213 \\
\texttt{ggordon@cs.cmu.edu} \\
}

\begin{document}
\maketitle
\begin{abstract}
We introduce the Reduced-Rank Hidden Markov Model (RR-HMM), a generalization of HMMs that can model smooth state evolution as in Linear Dynamical Systems (LDSs) as well as non-log-concave predictive distributions as in continuous-observation HMMs. RR-HMMs assume an $m$-dimensional latent state and $n$ discrete observations, with a transition matrix of rank $k \leq m$. This implies the dynamics evolve in a $k$-dimensional subspace, while the shape of the set of predictive distributions is determined by $m$. Latent state belief is represented with a $k$-dimensional state vector and inference is carried out entirely in $\Rbbvec{k}$, making RR-HMMs as computationally efficient as $k$-state HMMs yet more expressive. To learn RR-HMMs, we relax the assumptions of a recently proposed spectral learning algorithm for HMMs~\cite{zhang09} and apply it to learn $k$-dimensional observable representations of rank-$k$ RR-HMMs. The algorithm is consistent and free of local optima, and we extend its performance guarantees to cover the RR-HMM case. We show how this algorithm can be used in conjunction with a kernel density estimator to efficiently model high-dimensional multivariate continuous data. We also relax the assumption that single observations are sufficient to disambiguate state, and extend the algorithm accordingly. Experiments on synthetic data and a toy video, as well as on a difficult robot vision modeling problem, yield accurate models that compare favorably with standard alternatives in simulation quality and prediction capability.
\end{abstract}
\section{Introduction}
Models of stochastic discrete-time dynamical systems have important applications in a wide range of fields. Hidden Markov Models (HMMs)~\cite{rbnr:attr} and  Gaussian Linear Dynamical Systems (LDSs)~\cite{ghahramani96} are two examples of \textit{latent variable models} of dynamical systems, which  assume that sequential data points are noisy, incomplete observations of a latent state that evolves over time. HMMs model this latent state as a discrete variable, and represent belief as a discrete distribution over states. LDSs on the other hand model the latent state as a set of real-valued variables, are restricted to linear transition and observation functions,  and employ a Gaussian belief distribution. The distributional assumptions of HMMs and LDSs also result in important differences in the evolution of their belief over time. The discrete state of HMMs is good for modeling systems with mutually exclusive states that can have completely different observation signatures. The joint predictive distribution over observations is allowed to be non-log-concave when predicting or simulating the future, leading to what we call \emph{competitive inhibition} between states (see Figure~\ref{RRHMM-fig-clockResults} below for an example). Competitive inhibition denotes the ability of a model's predictive distribution to place probability mass on observations while disallowing mixtures of those observations. Conversely, the Gaussian joint predictive distribution over observations in LDSs is log-concave, and thus does not exhibit competitive inhibition. However, LDSs naturally model\emph{ smooth state evolution}, which  HMMs are particularly bad at. The dichotomy between the two models hinders our ability to compactly model systems that exhibit \emph{both} competitive inhibition and smooth state evolution.

We present the Reduced-Rank Hidden Markov Model (RR-HMM), a smoothly evolving dynamical model with the ability to represent nonconvex predictive distributions by relating discrete-state and continuous-state models. HMMs can approximate smooth state evolution by tiling the state space with a very large number of low-observation-variance discrete states with a specific transition structure. However, inference and learning in such a model is highly inefficient due to the large number of parameters, and due to the fact that existing HMM learning algorithms, such as Expectation Maximization (EM)~\cite{rbnr:attr}, are prone to local minima. RR-HMMs allow us to reap many of the benefits of large-state-space HMMs without incurring the associated inefficiency during inference and learning. Indeed, we show that all inference operations in the RR-HMM can be carried out in the low-dimensional space where the dynamics evolve, decoupling their computational cost from the number of hidden states. This makes \textit{rank}-$k$ RR-HMMs (with any number of states) as computationally efficient as $k$-\textit{state} HMMs, but much more expressive. Though the RR-HMM is in itself novel, its low-dimensional $\Rbbvec{k}$ representation is related to existing models such as Predictive State Representations (PSRs)~\cite{littman02}, Observable Operator Models (OOMs)~\cite{jaeger00a}, generalized HMMs~\cite{balasubramanian93}, and weighted automata~\cite{schutzenberger61,fleiss74}, as well as the the representation of LDSs learned using Subspace Identification~\cite{vanoverschee96book}. These and other related models and algorithms are discussed further in Section~\ref{RRHMM-sec-relwork}.

To learn RR-HMMs from data, we adapt a recently proposed spectral learning algorithm by Hsu, Kakade and Zhang~\cite{zhang09} (henceforth referred to as HKZ) that learns \textit{observable representations} of HMMs using matrix decomposition  and regression on empirically estimated observation probability matrices of past and future observations. An observable representation of an HMM allows us to model sequences with a series of operators without knowing the underlying stochastic transition and observation matrices. The HKZ algorithm is free of local optima and asymptotically unbiased, with a finite-sample bound on $L_1$ error in joint probability estimates
from the resulting model. However, the original algorithm and its bounds assume (1) that the transition model is full-rank and (2) that single observations are informative about the entire latent state, i.e.\@ \textit{$1$-step observability}. We show how to generalize the HKZ bounds to the low-rank transition matrix case and derive tighter bounds that depend on $k$ instead of $m$, allowing us to learn rank-$k$ RR-HMMs of arbitrarily large $m$ in $\mathcal{O}(Nk^2)$ time, where $N$ is the number of samples. We also describe and test a method for circumventing the $1$-step observability condition by combining observations to make them more informative. A version of this learning algorithm can learn general PSRs~\cite{boots09arxiv} though our error bounds don't yet generalize to this case.

Experiments show that our learning algorithm can recover the underlying RR-HMM in a variety of synthetic domains. We also demonstrate that RR-HMMs are able to compactly model smooth evolution \textit{and} competitive inhibition in a clock pendulum video, as well as in real-world mobile robot vision data captured in an office building. Robot vision data (and, in fact, most real-world multivariate time series data) exhibits smoothly evolving dynamics requiring multimodal predictive beliefs, for which RR-HMMs are particularly suited. We compare performance of RR-HMMs to LDSs and HMMs on simulation and prediction tasks. Proofs and details regarding examples are in the Appendix.
\section{Reduced-rank Hidden Markov Models}\label{RRHMM-sec-themodel}
\begin{figure}[!tb]
\begin{center}
\begin{tabular}{c}
\includegraphics[width=1\linewidth]{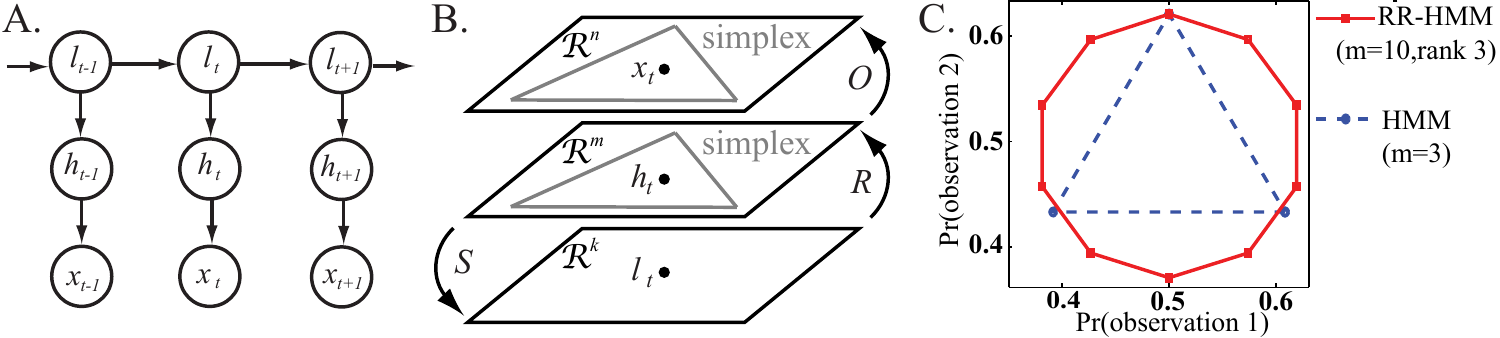}
\end{tabular}
\end{center}
\caption{ (A) The graphical model representation of an RR-HMM. $l_t$ denotes the $k$-dimensional state vector, $h_t$ the $m$-dimensional discrete state, and $x_t$ the discrete observation. The distributions over $h_t$ and $l_{t+1}$ are deterministic functions of $l_t$. (B) An illustration of different RR-HMM parameters and the spaces and random variables they act on. (C) Projection of sets of predictive distributions of a rank $3$ RR-HMM with $10$ states, and a $3$-state full-rank HMM with similar parameters.} \label{RRHMM-fig-diagram}
\end{figure}
Let $h_t \in {1,\ldots,m}$ denote the discrete hidden states of an HMM  at time $t$, and $x_t \in {1,\ldots,n}$ denote the discrete observations. Let $T \in \Rbbmat{m}{m}$ be the state transition probability matrix with $T_{ij} = \Pr[h_{t+1} = i \mid h_t = j]$. $O \in \Rbbmat{n}{m}$ is the observation probability matrix such that $O_{ij} = \Pr[x_t = i \mid h_t = j]$, and $\vec{\pi} \in \Rbbvec{m}$ is the initial state distribution with $\vec{\pi}_i = \Pr[h_1 = i]$. Let $\vec{h}_t \in \Rbbvec{m}$ denote the system's \textit{belief}, i.e.\@ a distribution over hidden states at time $t$. If we use $e_i$ to denote the $i^{th}$ column of the identity matrix, then $\vec{h}_t$ is equivalent to the conditional expectation of $e_{h_t}$, with the conditioning variables clear from context.
%
%
In addition to the standard HMM notation, assume $T$ has rank $k$ and let $T = R S$ where $R \in \Rbbmat{m}{k}$ and $S \in \Rbbmat{k}{m}$. This implies that the dynamics of the system can be expressed in $\Rbbvec{k}$ rather than $\Rbbvec{m}$. By convention, we think of $S$ as projecting the $m$-dimensional state distribution vector to a $k$-dimensional state vector, and $R$ as expanding this low-dimensional state back to an $m$-dimensional state distribution vector while propagating it forward in time. One possible choice for $R$ and $S$ is to use any $k$ independent columns of $T$ as the columns of $R$, and let the columns of $S$ contain the coefficients required to reconstruct $T$ from $R$, though other choices are possible (e.g.\@ using SVD). Also assume for now that $m \le n$ (we relax this assumption in Section~\ref{RRHMM-sec-stacking}). We denote the $k$-dimensional projection of the hidden state vector $\vec{h}_t$ as $\vec{l}_t$, which is simply a vector of real numbers rather than a stochastic vector. We assume the initial state distribution lies in the low dimensional space as well, i.e.\@ $\vec{\pi} = R \vec{\pi}_l$ for some vector $\vec{\pi}_l \in \Rbbvec{k}$. \@Figure~\ref{RRHMM-fig-diagram}(A) illustrates the graphical model corresponding to an RR-HMM. Figure~\ref{RRHMM-fig-diagram}(B) illustrates some of the different RR-HMM parameters and the spaces they act on.

To see how the probability of a sequence can be computed using these parameters, define
\[A_x = R S \diag(O_{x,1},...,O_{x,m})\]
so that $A_x \in \mathbb{R}^{m \times m}$, and define
\[W_x = S \diag(O_{x,1},...,O_{x,m}) R\]
so that $W_x \in \mathbb{R}^{k\times k}$. Also let $W = \sum_x W_x = S R$. With these definitions, the joint probability of $x_1,...,x_t$, can written using either$\{A_x\}$ or $\{W_x\}$  as
\begin{subequations} \label{eq:jointprobcalc}
\begin{align}
\Pr[x_1,...,x_t] &= \onevect_m   A_{x_t} \ldots A_{x_1} \pivec \\
&= \onevect_m  R W_{x_t} \ldots W_{x_1} \pivec_l
\end{align}
\end{subequations}
The latter parametrization casts a rank-$k$ RR-HMM as a $k$-dimensional PSR or transformed PSR~\cite{rosencrantz04}. Inference can be carried out in $\mathcal{O}(Nk^2)$ time in this representation. However, since every HMM is trivially a PSR, this leads to the question of how expressive rank-$k$ RR-HMMs are in comparison to $k$-state full-rank HMMs. The following example is instructive.
\subsection{Expressivity of RR-HMMs}
We describe a rank-$k$ RR-HMM whose set of possible predictive distributions is easy to visualize and describe. Consider the following rank $3$ RR-HMM with $10$ states and $4$ observations. The observation probabilities in each state are of the form
\begin{align*}
O_{i,\cdot}\ &=\ [\ p_iq_i\ \ \ p_i(1-q_i)\ \ \ (1-p_i)q_i\ \ \ (1-p_i)(1-q_i)]
\end{align*}
for some $0 \leq p_i,q_i \leq 1$, which can be interpreted as $4$ discrete observations, factored as two binary components which are independent given the state. $T$ and $p_i,q_i$ are chosen to place the vertices of the set of possible predictive distributions on evenly spaced points along a circle in $(p,q)$-space:
\begin{align*}
T_{ij} &= (1/2m)\left[2 + \sin\left(2\pi i/m\right)\sin\left(2\pi j/m\right) + \cos\left(2\pi i/m\right)\cos\left(2\pi j/m\right)\right] \\
p_i &= \left[\sin(2\pi i/m)+1\right]/2 \\
q_i &= \left[\cos(2\pi i/m)+1\right]/2
\end{align*}
We plot the marginal probability of each component of the observation, ranging across all achievable values of the latent state vector for the $m=10$ case (Figure~\ref{RRHMM-fig-diagram}(C)), yielding a $10$-sided polygon as the projection of the set of possible predictive distributions. These distributions are the columns of $T\tps O$. We also plot the corresponding marginals for the $m=3$ full-rank HMM case to yield a triangular set. More generally, from a $k$-state HMM, we can get at most a $k$-sided polygon for the set of possible predictive distributions.

The above example illustrates that rank-$k$ RR-HMMs with $m$ states can model sets of predictive distributions which full-rank HMMs with less than $m$ states cannot express. However, as we shall see, inference in rank-$k$ RR-HMMs of arbitrary $m$ is as efficient as inference in $k$-state full-rank HMMs. This implies that the additional degrees of freedom in the RR-HMM's low-dimensional parameters and state vectors buy it considerable expressive power. Since RR-HMMs are also related to PSRs as pointed out in the previous section, and since our learning algorithm will be shown to be \textit{consistent} for estimating PSRs (though we have finite-sample guarantees only for the RR-HMM case), it is also instructive to examine the expressivity of PSRs in general. We refer the reader to Jaeger (2000)~\cite{jaeger00a} and James et.\@ al.\@ (2004)~\cite{james04} for more on this.
%
%
\section{Learning Reduced-Rank HMMs}
\label{RRHMM-sec-learning} In a full-rank HMM, the maximum likelihood solution for the parameters $\{ T, O\} $ can be found through iterative techniques such as expectation maximization (EM)~\cite{baum}. EM, however, is prone to local optima and does not address the model selection problem. HMM model selection algorithms that avoid local minima (e.g.\@~\cite{siddiqi07a}) are better but still not guaranteed to return anything close to optimal as data increases, and is slow beyond a certain state space magnitude. Moreover, in learning RR-HMMs we face the additional challenge of learning the factors of its low-rank transition matrix. We could use EM to estimate $T$ followed by (or combined with) matrix factorization algorithms such as Singular Value Decomposition (SVD)~\cite{horn85book} or Non-negative Matrix Factorization (NMF)~\cite{hoyer04}. This approach has several drawbacks. For example, if the noisy estimate of a low-rank transition matrix is not low-rank itself, SVD could cause negative numbers to appear in the reconstructed transition matrix. Also, algorithms for NMF are only locally optimal, and NMF is overly restrictive in that it constrains its factor matrices to be non-negative, which is unnecessary for our application since low-rank transition matrices may have negative numbers in their factors $R$ and $S$.

An alternative approach, which we adopt, is to learn an asymptotically unbiased observable representation of an RR-HMM directly using SVD of a probability matrix relating past and future observations. This idea has roots in subspace identification~\cite{vanoverschee96book,katayama05} and multiplicity automata~\cite{schutzenberger61,fleiss74,balasubramanian93} as well as the PSR/OOM literature~\cite{jaeger00a,singh04} and was recently formulated in a paper by Hsu, Kakade and Zhang~\cite{zhang09} for full-rank HMMs. We use their algorithm, extending its theoretical guarantees for the low-rank HMM case where the rank of the transition matrix T is $k \leq m$. Computationally, the only difference in our base algorithm (before Section~\ref{RRHMM-sec-stacking}) is that we learn a rank $k$ representation instead of rank $m$. This allows us learn much more compact representations of possibly large-state-space real-world HMMs, and greatly increases the applicability of the original algorithm. Even when the underlying HMM is not low-rank, we can examine the singular values to tune the complexity of the underlying RR-HMM, thus providing a natural method for model selection. We present the main definitions, the algorithm and its performance bounds below. Detailed versions of the supporting proofs and lemmas can be found in the Appendix.
%
%
\subsection{The Algorithm}
\label{RRHMM-sec-learning-algo}
The learning algorithm depends on the following vector and matrix quantities that comprise properties of single observations, pairs of observations and triples:
\begin{align*}
[P_1]_i &= \Pr[x_1 = i] \\
[P_{2,1}]_{i,j} &= \Pr[x_2 = i, x_1 = j] \\
[P_{3,x,1}] &= \Pr[x_3 = i, x_2 = x, x_1 = j] \mbox{ for $x = 1,\ldots,n$}
\end{align*}
$P_1 \in \mathbb{R}^n$ is a vector, $P_{2,1} \in \mathbb{R}^{n \times n}$ and $P_{3,x,1} \in \mathbb{R}^{n\times n}$ are matrices.
 These quantities are closely related to matrices computed in algorithms for learning OOMs~\cite{jaeger00a}, PSRs~\cite{singh04} and LDSs using subspace identification (Subspace ID)~\cite{vanoverschee96book}. They can be expressed in terms of HMM parameters (for proofs see the Appendix: Lemmas~\ref{lem_HKZ_2} and~\ref{lem_HKZ_3} in Section~\ref{RRHMM-sec-prelims}):
\begin{align*}
\Pone\tps &= {\onevec_m}\tps T \diag(\pi)O\tps \\
\Ptwoone &= O T \diag(\pi)O\tps \\
\Pthreexone &= O A_x T \diag(\pi)O\tps
\end{align*}
Note that $\Ptwoone$ and $\Pthreexone$ both contain a factor of $T$ and hence are both of rank $k$ for a rank-$k$ RR-HMM. This property will be important for recovering an estimate of the RR-HMM parameters from these matrices. The primary intuition is that, when projected onto an appropriate linear subspace,  $\Pthreexone$ is linearly related to $\Ptwoone$ through a product of RR-HMM parameters. This allows us to devise an algorithm that
\begin{enumerate}
\item estimates $\Ptwoone$ and $\Pthreexone$ from data,
\item projects them to an appropriate linear subspace computed using SVD,
\item uses linear regression to estimate the RR-HMM parameters (up to a similarity transform) from these projections.
\end{enumerate}

Specifically, the algorithm attempts to learn an \textit{observable representation} of the RR-HMM using a matrix  $U\in\Rbbmat{n}{k}$ such that $U\tps OR$ is invertible. An observable representation is defined as follows.
\begin{definition} \label{def_observable_rep} The \underline{observable representation} is defined to be the parameters $b_1,b_\infty,\{B_x\}_{x=1}^n$ such that:
\begin{subequations}
\begin{align}
\vec{b}_1 &= U\tps  P_1 \\
\vec{b}_\infty &= (P_{2,1}\tps  U)^+ P_1 \\
B_x &= (U\tps  P_{3,x,1})(U\tps  P_{2,1})^+ \mbox{\ \ \ \ \ for $x = 1,\ldots,n$}
\end{align}
\end{subequations}
\end{definition}
For the RR-HMM, note that the dimensionality of the parameters is determined by $k$, not $m$: $b_1 \in \mathbb{R}^{k}$, $b_\infty \in \mathbb{R}^{k}$ and $\forall x \ \ B_x \in \mathbb{R}^{k\times k}$. Though these definitions seem arbitrary at first sight, the observable representation of the RR-HMM is closely related to the true parameters of the RR-HMM in the following manner (see Lemma~\ref{lem_HKZ_3} in the Appendix for the proof):
\begin{enumerate}
\item $\bone = \UOR\pi_l  = \UO\pi$,
\item $\binft = 1_m\tps  R \UORi$,
\item For all $x = 1,\ldots,n:\ B_x =\UOR W_x \UORi$
\end{enumerate}
Hence $B_x$ is a similarity transform of the RR-HMM parameter matrix $W_x = S\diag(O_{x,\cdot})R$ (which, as we saw earlier, allows us to perform RR-HMM inference), and $\bone$ and $\binf$ are the corresponding linear transformations of the RR-HMM initial state distribution and the RR-HMM normalization vector. Note that $\UOR$ must be invertible for these relationships to hold. Together, the parameters $\bone$,$\binf$ and $B_x$ for all $x$ comprise the observable representation of the RR-HMM. Our learning algorithm will estimate these parameters from data. The algorithm for estimating rank-$k$ RR-HMMs is equivalent to the spectral HMM learning algorithm of HKZ~\cite{zhang09} for learning $k$-state HMMs. Our relaxation of their conditions (e.g.\@ HKZ assume a full-rank transition matrix, without which their bounds are vacuous), and our performance guarantees for learning rank-$k$ RR-HMMs, show that the algorithm learns a much larger class of $k$-dimensional models than the class of $k$-state HMMs.

\paragraph{\textsc{Learn-RR-HMM}($k,N$)} The learning algorithm takes as input the desired rank $k$ of the underlying RR-HMM rather than the number of states $m$. Alternatively, given a singular value threshold the algorithm can choose the rank of the HMM by examining the singular values of $\Ptwoone$ in Step 2. It assumes that we are given $N$ independently sampled observation triples $(x_1,x_2,x_3)$ from the HMM\@. In practice, we can use a single long sequence of observations as long as we discount the bound on the number of samples based on the mixing rate of the HMM (i.e.\@ ($1$ $-$ the second eigenvalue of $T$)), in which case $\pi$ must correspond to the stationary distribution of the HMM to allow estimation of $\Pone$.  The algorithm results in an \textit{estimated observable representation} of the RR-HMM, with parameters $\widehat{b}_1, \widehat{b}_\infty$, and $\widehat{B}_x$ for $x = 1,\ldots,n$. The steps are briefly summarized here for reference:
\begin{enumerate}
\item Compute empirical estimates $\Phone,\Phtwoone,\Phthreexone$ of $\Pone,\Ptwoone,\Pthreexone$ (for $x = 1,...,n$).
\item Use SVD on $\Phtwoone$ to compute $\Uhat$, the matrix of left singular vectors corresponding to the $k$ largest singular values.
\item Compute model parameter estimates:
    \begin{enumerate}
    \item $\bhat_1 = \Uhat\tps\Phone$,
    \item $\bhat_\infty = (\Phtwoone\tps\Uhat)^+\Phone$,
    \item $\Bhatx = \Uhat\tps\Phthreexone(\Uhat\tps\Phtwoone)^+$ (for $x = 1,\ldots,n$)
    \end{enumerate}
\end{enumerate}
We now examine how we can perform inference in the RR-HMM using the observable representation. For this, we will need to define the \textit{internal state} $\bvec_t$. Just as the parameter $\bone$ is a linear transform of the initial RR-HMM belief state, $\bvec_t$ is a linear transform of the belief state of the RR-HMM at time $t$ (Lemma~\ref{lem_HKZ_4} in Section~\ref{RRHMM-sec-prelims} of the Appendix):
\begin{align*}
\bvec_t &= \UOR \lvec_t(x_{1:t-1}) = \UO \hvec_t(x_{1:t-1})
\end{align*}
This internal state $\bvec_t$ can be updated to condition on observations and evolve over time, just as we can update $\lvec_t$ for RR-HMMs and $\hvec_t$ for regular HMMs.
\subsection{Inference in the Observable Representation}
Given a set of observable parameters, we can predict the probability of a sequence, update the internal state $\widehat{b}_t$ to perform filtering and predict conditional probabilities as follows (see Lemma~\ref{lem_HKZ_4} in the Appendix for proof):
\begin{itemize}
\item Predict sequence probability: $ \widehat{\Pr}[x_1,\ldots,x_t] = \bhinft\widehat{B}_{x_t} \ldots \widehat{B}_{x_1} \bhone $
\item Internal state update: $ \widehat{b}_{t+1} = \frac{\widehat{B}_{x_t} \bhat_t}{\bhinft \widehat{B}_{x_t} \bhat_t} $
\item Conditional probability of $x_t$ given $x_{1:{t-1}}$: $ \widehat{\Pr}[x_t \mid x_{1:t-1}] = \frac{\bhinft \widehat{B}_{x_t} \bhat_t}{\sum_x \bhinft \widehat{B}_{x} \bhat_t} $
\end{itemize}

Estimated parameters can, in theory, lead to negative probability estimates.  These are most harmful when they cause the normalizers $\widehat{b}\tps_\infty \widehat{B}_{x_t} \widehat{b}_t$ or $\sum_x \widehat{b}\tps_\infty \widehat{B}_{x} \widehat{b}_t$ to be negative. However, in our experiments, the latter was never negative and the former was very rarely negative; and, using real-valued observations with KDE (as in Section~\ref{RRHMM-sec-realobs}) makes negative normalizers even less likely, since in this case the normalizer is a weighted sum of several estimated probabilities.  In practice we recommend thresholding the normalizers with a small positive number, and not trusting probability estimates for a few steps if the normalizers fall below the threshold.

Note that the inference operations occur entirely in $\Rbbvec{k}$. We mentioned earlier that parameterizing RR-HMM parameters as $W_x$ for all observations $x$ casts it as a PSR of $k$ dimensions. In fact the learning and inference algorithms for RR-HMMs proposed here have no dependence on the number of states $m$ whatsoever, though other learning algorithms for RR-HMMs can depend on $m$ (e.g.\@ if they learn $R$ and $S$ directly). The RR-HMM formulation is intuitively appealing due to the idea of a large discrete state space with low-rank transitions, but this approach is also a provably \textit{consistent learning algorithm for PSRs} in general, with finite-sample performance guarantees for the case where the PSR is an RR-HMM. Since PSRs are provably more expressive and compact than finite-state HMMs~\cite{jaeger00a,james04}, this indicates that we can learn a more powerful class of models than HMMs using this algorithm.
%
\subsection{Theoretical Guarantees}
The following finite sample bound on the estimated model generalizes analogous results from HKZ to the case of low-rank $T$. Theorem~\ref{thm_HKZ_6} bounds the $L_1$ error in joint probability estimates from the learned model. This bound shows the consistency of the algorithm in learning a correct observable representation of the underlying RR-HMM, without ever needing to recover the high-dimensional parameters $R,S,O$ of the latent representation. Note that our error bounds are \emph{independent} of $m$, the number of hidden states; instead, they depend on $k$, the rank of the transition matrix, which can be much smaller than $m$. Since HKZ explicitly assumes a full-rank HMM transition matrix, and their bounds become vacuous otherwise, generalizing their framework involves relaxing this condition, generalizing the theoretical guarantees of HKZ and deriving proofs for these guarantees.

Define $\sigma_k(M)$ to denote the $k^{\mathrm{th}}$ largest singular value of matrix $M$. The sample complexity bounds depend polynomially on $1/\sigma_k (P_{2,1})$ and $1/\sigma_k(OR)$. The larger $\sigma_k (P_{2,1})$ is, the more well-separated are the dynamics from noise. The larger $\sigma_k(OR)$ is, the more informative the observation is regarding state. For both these quantities, the larger the magnitude, the fewer samples we need to learn a good model. The bounds also depend on a term $n_0(\epsilon)$, which is the minimum number of observations that account for $(1-\epsilon)$ of the total probability mass, i.e.\@ the number of ``important'' observations. Recall that $N$ is the number of independently sampled observation triples which comprise the training data, though as mentioned earlier we can also learn from a single long training sequence.

The theorem holds under mild conditions. Some of these are the same as (or relaxations of) conditions in HKZ, namely that the prior $\pivec$ is nonzero everywhere, and a number of matrices of interest $(R,S,O,\UOR)$ are of rank at least $k$ for invertibility reasons. The other conditions are unique to the low-rank setting, namely that $S\diag(\pivec)O\tps$ has rank at least $k$, $R$ has at least one column whose $L_2$ norm is at most $\sqrt{k/m}$, and the $L_1$ norm of $R$ is at most $1$. The first of these conditions implies that the column space of $S$ and the row space of $O$ have some degree of overlap. The other two are satisfied, in the case of HMMs, by thinking of $R$ as containing $k$ linearly independent probability distributions along its columns (including a near-uniform column) and of $S$ as containing the coefficients needed to obtain $T$ from those columns. Alternatively, the conditions can be satisfied for an arbitrary $R$ by scaling down entries of $R$ and scaling up entries of $S$ accordingly. However, this increases $1/\sigma_k(OR)$, and hence we pay a price by increasing the number of samples needed to attain a particular error bound. See the Appendix (Section~\ref{RRHMM-sec-prelims}) for formal statements of these conditions.

\begin{theorem} \label{thm_HKZ_6}[Generalization of HKZ Theorem 6] There exists a constant $C > 0$ such that the following holds. Pick any $0 \le \epsilon,\eta \le 1$ and $t \geq 1$. Assume the HMM obeys Conditions 3,4,5,6 and 7. Let $\varepsilon = \sigma_k(OR) \sigma_k(\Ptwoone)\epsilon/(4t\sqrt{k})$. Assume\\
\[ N \geq C \cdot \frac{t^2}{\epsilon^2} \cdot \left(\frac{k}{\sigma_k(OR)^2 \sigma_k(P_{2,1})^4} + \frac{k\cdot n_0(\varepsilon)}{\sigma_k(OR)^2 \sigma_k (P_{2,1})^2}  \right)\cdot \log (1/\eta) \]
With probability $\geq 1 - \eta$, the model returned by \textsc{LearnRR-HMM}($k,N$) satisfies
\[ \sum_{x_1,\ldots,x_t} |\Pr[x_1,\ldots,x_t] - \widehat{\Pr}[x_1,\ldots,x_t]| \leq \epsilon \]
\end{theorem}
For the proof, see the Appendix (Section~\ref{RRHMM-sec-proofs-proof1}).
%
%
\subsection{Learning with Observation Sequences as Features}
\label{RRHMM-sec-stacking} The probability matrix $\Ptwoone$ relates one past timestep to one future timestep, under the assumption that the vector of observation probabilities at a single step is sufficient to disambiguate state ($n \geq m$ and $\rank(O) = m$). In system identification theory, this corresponds to assuming \textit{$1$-step observability}~\cite{vanoverschee96book}. This assumption is unduly restrictive for many real-world dynamical systems of interest. More complex sufficient statistics of past and future may need to be modeled, such as the \textit{block Hankel matrix} formulations for subspace methods~\cite{vanoverschee96book,katayama05} to identify linear systems that are not $1$-step observable.

For RR-HMMs, this corresponds to the case where $n < m$ and/or $\rank(O) < m$. Similar to the Hankel matrix formulation, we can stack multiple observation vectors such that each augmented observation comprises data from several, possibly consecutive, timesteps. The observations in the augmented observation vectors are assumed to be non-overlapping, i.e.\@ all observations in the new observation vector at time $t+1$ have larger time indices than observations in the new observation vector at time $t$. This corresponds to assuming \textit{past sequences} and \textit{future sequences} spanning multiple timesteps as events that characterize the dynamical system, causing $\Pone$,$\Ptwoone$ and $\Pthreexone$ to be larger. Note that the $x$ in $\Pthreexone$ still denotes a \textit{single} observation, whereas the other indices in $\Pone$, $\Ptwoone$ and $\Pthreexone$ are now associated with events. For example, if we stack $\overline{n}$ consecutive observations, $\Pthreexone[i,j]$ equals the probability of seeing the $i^{\mathrm{th}}$ $\overline{n}$-length sequence, followed by the single observation $x$, followed by the $j^{th}$ $\overline{n}$-length sequence. Empirically estimating this matrix consists of scanning for the appropriate subsequences $i$ and $j$ separated by observation symbol $x$, and normalizing to obtain the occurrence probability.

$\Ptwoone$ and $\Pthreexone$ become larger matrices if we use a larger set of events in the past and future. However, stacking observations does not complicate the \textit{dynamics}: it can be shown that the rank of $\Ptwoone$ and $\Pthreexone$ cannot exceed $k$ (see Section~\ref{APPENDIX-sec-ambig} in the Appendix for a proof sketch). Since our learning algorithm relies on an SVD of $\Ptwoone$, this means that augmenting the observations does not increase the rank of the HMM we are trying to recover. Also, since $\Pthreexone$ is still an observation probability matrix with respect to a \textit{single} unstacked observation $x$ in the middle, the number of observable operators we need remains constant. Our complexity bounds successfully generalize to this case, since they only rely on $\Pone$, $\Ptwoone$ and $\Pthreexone$ being matrices of probabilities summing to $1$ (for the former two) or to $\Pr[x_2 = x]$ (for the latter), as they are here.

The extension given above for learning HMMs with ambiguous observations differs from the approach suggested by HKZ, which simply substitutes observations with \textit{overlapping} tuples of observations (e.g.  $\overline{\Ptwoone}(j,i) = \Pr[x_3 = j_2, x_2 = j_1, x_2 = i_2, x_1 = i_1]$). There are two potential problems with the HKZ approach. First,  the number of observable operators increases exponentially with the length of each tuple: there is one observable operator per tuple, instead of one per observation. Second, $\overline{\Ptwoone}$ cannot be decomposed into a product of matrices that includes $T$, and consequently no longer has rank equal to the rank of the HMM being modeled. Thus, the learning algorithm could require much more data to recover a correct model if we use the HKZ approach.
\subsection{Learning with Real-Valued Observations}
%
\label{RRHMM-sec-realobs} The default RR-HMM formulation assumes discrete observations. However, since the model formulation converts the discrete observations into $n$-dimensional probability vectors, and the filtering, smoothing and learning algorithms we discuss all do the same, it is straightforward to model multivariate continuous data with Kernel Density Estimation~\cite{silverman:1986}.

This affects the learning algorithm and inference procedure as follows. Assume for ease of notation that the training data consists of $N$ sets of three consecutive continuous observation vectors each, i.e., $\left\{\langle\xvec_{1,1},\xvec_{1,2},\xvec_{1,3}\rangle,\langle\xvec_{2,1},\xvec_{2,2},\xvec_{2,3}\rangle,\ldots,\langle\xvec_{N,1},\xvec_{N,2},\xvec_{N,3}\rangle\right\}$, though in practice we could be learning from a single long sequence (or several). Also assume for now that each observation vector contains a single raw observation, though this technique can easily be combined with the more sophisticated sequence-based learning and feature-based learning methods described above. Pick a kernel function $K(\cdot)$ and $n$ kernel centers $\cvec_1\ldots\cvec_n$. (In general we can use different kernels and centers for different feature vectors.) Let $\lambda$ be a bandwidth parameter that goes to zero at the appropriate rate in the limit.

First compute $n\times 1$ feature vectors $\langle\phivec_j\rangle_{j=1}^N$, $\langle\psivec_j\rangle_{j=1}^N$, $\langle\xivec_j\rangle_{j=1}^N$ and $\langle\zetavec_j\rangle_{j=1}^N$, and normalize each to sum to 1:
\begin{align*}
[\phivec_j]_i &\propto K(\xvec_{j,1} - \cvec_i) \ \ \ \ [\psivec_j]_i \propto K(\xvec_{j,2} - \cvec_i) \\
[\xivec_j]_i &\propto  K(\xvec_{j,3} - \cvec_i) \ \ \ \ [\zetavec_j]_i \propto K\left((\xvec_{j,2} - \cvec_i)/\lambda\right)
\end{align*}
Note that for the second observation (in $\langle\zetavec_j\rangle_{j=1}^N$) we scale the kernel function by the bandwidth. Then, estimate the vector $\Pone$ and matrices $\Ptwoone$ and $\Pthreexone$ (for $\xvec = \cvec_1,\ldots,\cvec_n$) from data:
\begin{align*}
\Phone &= \frac{1}{N} \sum_{j=1}^N \phivec_j \ \ \ \ \ \Phtwoone = \frac{1}{N}\sum_{j=1}^N \psivec_j \phivec_j\tps \\
\mbox{For\ } x &= c_1,\ldots,c_n \mbox{:\ \ }\Phthreexone = \frac{1}{N}\sum_{j=1}^N [\zetavec_j]_x\xivec_j \phivec_j\tps
\end{align*}
We compute $n$ `base' observable operators $B_{c_1},\ldots,B_{c_n}$ from the estimated probability matrices, as well as vectors $\bone$ and $\binf$, using algorithm \textsc{Learn-RR-HMM} (Section~\ref{RRHMM-sec-learning-algo}). Given these parameters, filtering for a sequence $\langle\xvec_1,\ldots,\xvec_\tau\rangle$ now proceeds as follows: 
\begin{align*}
\mbox{For\ } t &= 1,\ldots,\tau \mbox{:} \\
&\mbox{Compute and normalize} [\sigmavec_{t}]_i \propto K\left((\xvec_t - \cvec_i)/\lambda\right). \\
& B_{\sigma_t} = \sum_{j=1}^n [\sigmavec_t]_j B_{c_j} \\
\vspace{-.05in} & \bvec_{t+1} = \frac{B_{\sigma_t}\bvec_t}{\binf B_{\sigma_t}\bvec_t}
\end{align*}
Our theoretical results carry over to the KDE case with modifications described in the RR-HMM document. Essentially, the bound still holds for predicting functions of $\sigmavec_1,\sigmavec_2,\ldots,\sigmavec_t$, though we do not yet have results connecting this bound to the error in estimating probabilities of raw observations. 
\section{Experimental Results}
\label{RRHMM-sec-results}
We designed several experiments to evaluate the properties of RR-HMMs and the learning algorithm both on synthetic and on real-world data. The first set of experiments (Section \ref{RRHMM-sec-results-synth}) tests the ability of the spectral learning algorithm to recover the correct RR-HMM\@. The second experiment (Section \ref{RRHMM-sec-results-clock}) evaluates the representational capacity of the RR-HMM by learning a model of a video that requires both competitive inhibition and smooth state evolution. The third set of experiments (Section \ref{RRHMM-sec-results-robot}) tests the model's ability to learn, filter, predict, and simulate video captured from a robot moving in an indoor office environment.
\subsection{Learning Synthetic RR-HMMs}
\label{RRHMM-sec-results-synth}
\begin{figure}[!tb]
\begin{tabular}{c}
\includegraphics[width=1\linewidth]{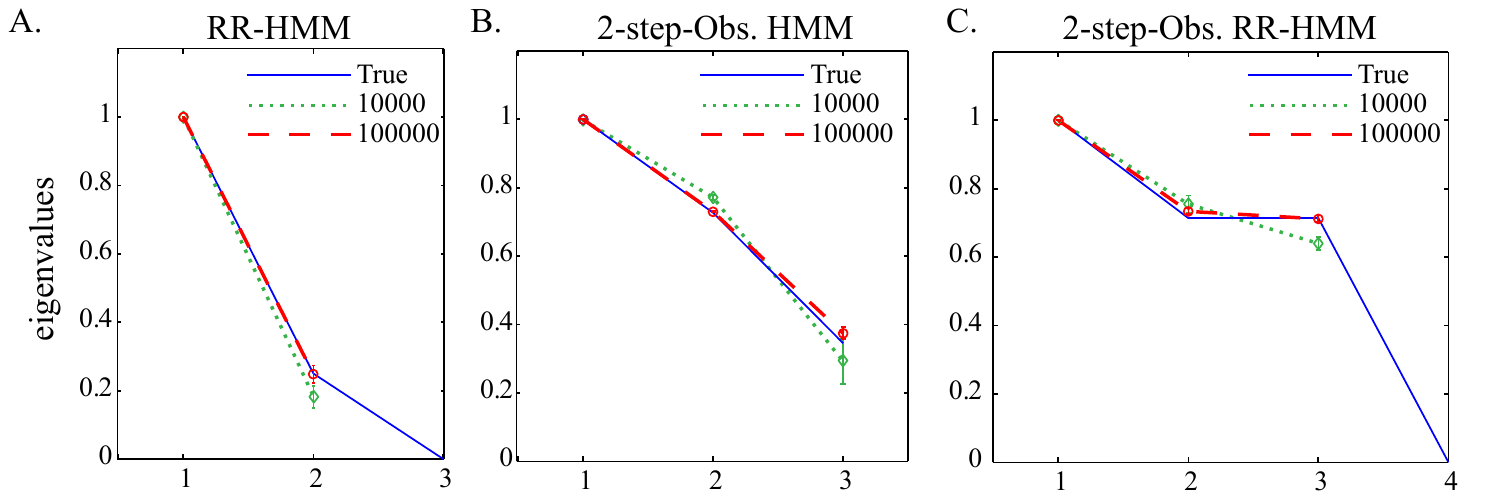}
\end{tabular}
\caption{Learning discrete RR-HMMs. The three figures depict the actual eigenvalues of three different RR-HMM transition matrices, and the eigenvalues (95\% error bars) of the sum of RR-HMM observable operators estimated with $10,000$ and $100,000$ training observations. (A) A 3-state, 3-observation, rank 2 RR-HMM. (B) A full-rank, 3-state, 2-observation HMM. (C) A 4-state, 2-observation, rank 3 RR-HMM.  } \label{RRHMM-fig-synthetic}
\end{figure}
First we evaluate the unbiasedness of the spectral learning algorithm for RR-HMMs on $3$ synthetic examples. In each case, we build an RR-HMM, sample observations from the model, and estimate the model with the spectral learning algorithm described in Section~\ref{RRHMM-sec-learning}. We compare the eigenvalues of $B = \sum_x B_x$ in the learned model to the eigenvalues of the transition matrix $T$ of the true model. $B$ is a similarity transform of $S\cdot R$ which therefore has the same non-zero eigenvalues as $T = R S$, so we expect the estimated eigenvalues to converge to the true eigenvalues with enough data. This is a necessary condition for unbiasedness but not a sufficient one. See Section~\ref{APPENDIX-sec-synth} in Appendix for parameters of HMMs used in the examples below.

\paragraph{Example 1: An RR-HMM} We examine an HMM with $m=3$ hidden states, $n=3$ observations, a full-rank observation matrix and a $k=2$ rank transition matrix. Figure~\ref{RRHMM-fig-synthetic}(A) plots the true and estimated eigenvalues for increasing size of dataset, along with error bars, suggesting that we recover the true dynamic model.
\paragraph{Example 2: A 2-step-Observable HMM} We examine an HMM with $m=3$ hidden states, $n=2$ observations, and a full-rank transition matrix (see Appendix for parameters). This HMM violates the $m \leq n$ condition. The parameters of this HMM cannot be estimated with the original learning algorithm, since a single observation does not provide enough information to disambiguate state. By stacking $2$ consecutive observations (see Section \ref{RRHMM-sec-stacking}), however,
the spectral learning algorithm can be applied successfully (Figure~\ref{RRHMM-fig-synthetic}(B)).
\paragraph{Example 3: A 2-step-Observable RR-HMM} We examine an HMM with $m=4$ hidden states, $n=2$ observations, and a $k=3$ rank transition matrix (see Appendix for parameters). In this example, the HMM is low rank \textit{and} multiple observations are required to disambiguate state. Again, stacking two consecutive observations in conjunction with the spectral learning algorithm is enough to recover good RR-HMM parameter estimates (Figure~\ref{RRHMM-fig-synthetic}(C)).
\subsection{Competitive Inhibition and Smooth State Evolution in Video}
\label{RRHMM-sec-results-clock}
\begin{figure}[!tb]
\begin{tabular}{c}
\includegraphics[width=1\linewidth]{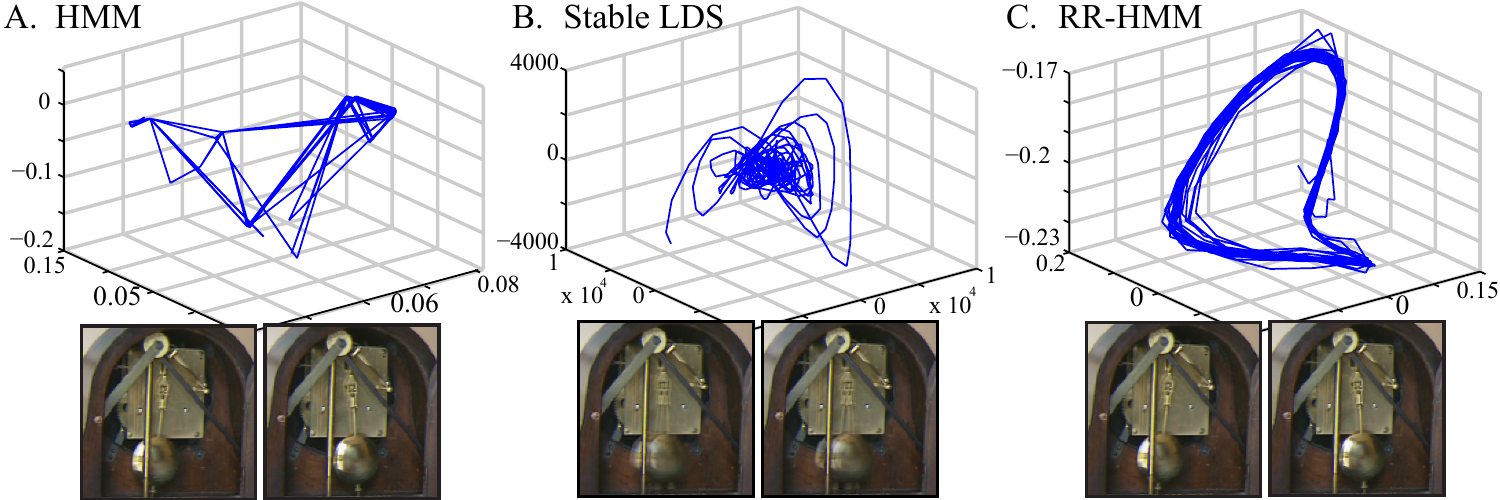}
\end{tabular}
\caption{The clock video texture simulated by a HMM, a stable LDS, and a RR-HMM. (A) The clock modeled by a $10$-state HMM. The manifold consists of the top $3$ principal components of predicted observations during simulation. The generated frames are coherent but motion in the video is jerky. (B) The clock modeled by a $10$-dimensional LDS. The manifold indicates the trajectory of the model in state space during simulation. Motion in the video is smooth but frames degenerate to superpositions. (C) The clock modeled by a rank $10$ RR-HMM. The manifold consists of the trajectory of the model in the low dimensional subspace of the state space during simulation. Both the motion and the frames are correct.} \label{RRHMM-fig-clockResults}
\end{figure}
We model a clock pendulum video consisting of 55 frames (with a period of $\sim 22$ frames) as a 10-state HMM, a 10-dimensional LDS, and a rank $10$ RR-HMM with $4$ stacked observations. Note that we could easily learn models with more than 10 latent states/dimensions; we limited the dimensionality in order to demonstrate the relative expressive power of the different models. For the HMM, we convert the continuous data to discrete observations by 1-NN on 25 kernel centers sampled sequentially from the training data. We trained the resulting discrete HMM using EM\@. We learned the LDS directly from the video using subspace ID with stability constraints~\cite{siddiqi07b} using a Hankel matrix of $10$ stacked observations. We trained the RR-HMM by stacking 4 observations, choosing an approximate rank of 10 dimensions, and learning 25 observable operators corresponding to 25 Gaussian kernel centers. We simulate a series of 500 observations from the model and compare the manifolds underlying the simulated observations and frames from the simulated videos  (Figure~\ref{RRHMM-fig-clockResults}). The small number of states in the HMM is not sufficient to capture the smooth evolution of the clock: the simulated video is characterized by realistic looking frames, but exhibits jerky irregular motion. For the LDS, although the $10$-dimensional subspace captures smooth evolution of the simulated video, the system quickly degenerates and individual frames of video are modeled poorly (resulting in superpositions of pendulums in generated frames). For the RR-HMM, the simulated video benefits from both smooth state evolution \emph{and} competitive inhibition. The manifold in the $10$-dimensional subspace is smooth and structured and the video is realistic. The results demonstrate that the RR-HMM has the benefits of smooth state evolution and compact state space of a LDS and the benefit of competitive inhibition of a HMM.
\subsection{Filtering, Prediction, and Simulation with Robot Vision Data}
\label{RRHMM-sec-results-robot}
\begin{figure}[tb!]
\begin{center}
\includegraphics[width=1\linewidth]{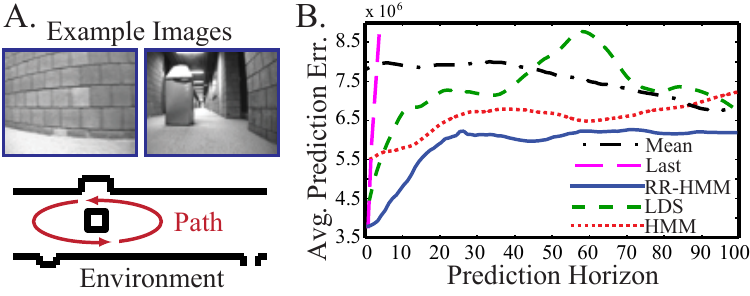}
\end{center}
\caption{(A) Sample images from the robot's camera. The figure below depicts the hallway environment with a central obstacle (black) and the path that the robot took through the environment while collecting data (the red counter-clockwise ellipse) (B) Squared error for prediction ($1,\ldots,100$ steps out in future) with different estimated models and baselines, averaged over different initial filtering durations ($1,\ldots,250$). } \label{RRHMM-fig-robotFig}
\end{figure}
We compare HMMs, LDSs, and RR-HMMs on the problem of modeling video data from a mobile robot in an indoor environment. A video of $2000$ frames was collected at $6$ Hz from a Point Grey Bumblebee2 stereo camera mounted on a Botrics Obot d100 mobile robot platform circling a stationary obstacle (Figure \ref{RRHMM-fig-robotFig}(A)) and $1500$ frames were used as training data for each model. Each frame from the training data was reduced to $100$ dimensions via SVD on single observations. Using this training data, we trained an RR-HMM ($k=50,n=1500)$ using spectral learning with sequences of 20 continuous observations (Section \ref{RRHMM-sec-stacking}) and KDE with Gaussian kernels (Section~\ref{RRHMM-sec-realobs}) with $1500$ centers, a 50-dimensional LDS using Subspace ID with Hankel matrices of $20$ timesteps, and a 50-state HMM with $1500$ discrete observations using EM run until convergence. For each model, we performed filtering for different extents $t_1=100,101,\ldots,250$, then predicted an image which was a further $t_2$ steps in the future, for $t_2 = 1,2\ldots,100$. The squared error of this prediction in pixel space was recorded, and averaged over all the different filtering extents $t_1$ to obtain means which are plotted in Figure~\ref{RRHMM-fig-robotFig}(B). As baselines, we plot the error obtained by using the mean of filtered data as a predictor (`Mean'), and the error obtained by using the last filtered observation (`Last').

Both baselines perform worse than any of the more complex algorithms (though as expected, the `Last' predictor is a good one-step predictor), indicating that this is a nontrivial prediction problem. The LDS does well initially (due to smoothness), and the HMM does well in the longer run (due to competitive inhibition), while the RR-HMM performs as well or better at both time scales since it models both the smooth state evolution and competitive inhibition in its predictive distribution. In particular, the RR-HMM yields significantly lower prediction error consistently for the duration of the prediction horizon (100 timesteps, i.e.\@ 16 seconds).
\section{Related Work}
\label{RRHMM-sec-relwork}
\subsection{Predictive State Representations}
Predictive State Representations (PSRs)~\cite{littman02,singh04} and Observable Operator Models (OOMs)~\cite{jaeger00a} model sequence probabilities as a product of \textit{observable operator} matrices. This idea, as well as the idea of learning such models using linear algebra techniques, originates in the literature on multiplicity automata and weighted automata ~\cite{schutzenberger61,fleiss74,balasubramanian93}. Despite recent improvements~\cite{wingate07a,zhao07}, practical learning algorithms for PSRs and OOMs have been lacking. RR-HMMs and its spectral learning algorithm are also closely related to methods in \textit{subspace identification}~\cite{vanoverschee96book,katayama05} in control systems for learning LDS parameters, which use SVD to determine the relationship between hidden states and observations.

As pointed out earlier, the spectral learning algorithm presented here learns PSRs. We briefly discuss other algorithms for learning PSRs from data. Several learning algorithms for PSRs have been proposed~\cite{singh03,james04,wolfe05}. It is easier for PSR learning algorithms to return \textit{consistent} parameter estimates because the parameters are based on observable quantities. \cite{rosencrantz04} develops an SVD-based method for finding a low-dimensional variant of PSRs, called \textit{Transformed PSRs} (TPSRs). Instead of tracking the probabilities of a small number of tests, TPSRs track a small number of linear combinations of a larger number of tests. This allows more compact representations, as well as dimensionality selection based on examining the singular values of the decomposed matrix, as in subspace identification methods. Note that nonlinearity can be encoded into the design of core tests. \cite{rudary03} introduced the concept of \textit{e-tests} in PSRs that are indicator functions of aggregate sets of future outcomes, e.g.\@ all sequence of observations in the immediate future that end with a particular observation after $k$ timesteps. In general, tests in discrete PSRs can be indicator functions of arbitrary statistics of future events, thus encoding nonlinearities that might be essential for modeling some dynamical systems. Recently, Exponential Family PSRs (EFPSRs)~\cite{wingate07a} were introduced as an attempt to generalize the PLG model to allow general exponential family distributions over the next $N$ observations. In the EFPSR, state is represented by modeling the parameters of a time-varying exponential family distribution over the next $N$ timesteps. This allows graphical structure to be encoded in the distribution, by choosing the parameters accordingly. The justification for choosing an exponential family comes from maximum entropy modeling. Though inference and parameter learning are difficult in graphical models of non-trivial structure, approximate inference methods can be utilized to make these problems tractable. Like PLGs, the dynamical component of EFPSRs is modeled by \textit{extending} and \textit{conditioning} the distribution over time. However, the method presented~\cite{wingate07a} has some drawbacks, e.g.\@ the extend-and-condition method is inconsistent with respect to marginals over individual timesteps between the extended and un-extended distributions.
\subsection{Hybrid Models, Mixture Models and other recent approaches}
RR-HMMs and their algorithms are also related to other \textit{hybrid models}. Note that previous models of the same name (e.g.\@~\cite{kumar98}) address a completely different problem, i.e.\@ reducing the rank of the Gaussian observation parameters. Since shortly after the advent of LDSs, there have been attempts to combine the discrete states of HMMs with the smooth dynamics of LDSs. We perform a brief review of the literature on hybrid models; see~\cite{ghahramani00variational} for a more thorough review. \cite{ackerson70} formulates a switching LDS variant where both the state and observation variable noise models are mixture of Gaussians with the mixture switching variable evolving according to Markovian dynamics, and derives the (intractable) optimal filtering equations where the number of Gaussians needed to represent the belief increases exponentially over time. They also propose an approximate filtering algorithm for this model based on a single Gaussian. \cite{shumway91} proposes learning algorithms for an LDS with switching observation matrices. \cite{bar-shalom93} reviews models where both the observations and state variable switch according to a discrete variable with Markov transitions. \textit{Hidden Filter HMMs} (HFHMMs)~\cite{fraser93} combine discrete and real-valued state variables and outputs that depend on both. The real-valued state is deterministically dependent on previous observations in a known manner, and only the discrete variable is hidden. This allows exact inference in this model to be tractable. \cite{chen00} formulates the \textit{Mixture Kalman Filter} (MKF) model along with a filtering algorithm, similar to~\cite{ackerson70} except that the filtering algorithm is based on sequential Monte-Carlo sampling.

The commonly used \textit{HMMs with mixture-model observations} (e.g.\@, Gaussian mixture) are a special case of RR-HMMs. A $k$-state HMM where each state corresponds to a Gaussian mixture of $m$ observation models of $n$ dimensions each is subsumed by a $k$-rank RR-HMM with $m$ distinct continuous observations of $n$ dimensions each, since the former is constrained to be non-negative and $\leq 1$ in various places (the $k$-dimensional transition matrix, the $k$-dimensional belief vector, the matrix which transforms this belief to observation probabilities) where the latter is not.

\textit{Switching State-Space Models} (SSSMs)~\cite{ghahramani00variational} posit the existence of several real-valued hidden state variables that evolve linearly, with a single Markovian discrete-valued switching variable selecting the state which explains the real-valued observation at every timestep. Since exact inference and learning are intractable in this model, the authors derive a structured variational approximation that decouples the state space and switching variable chains, effectively resulting in Kalman smoothing on the state space variables and HMM forward-backward on the switching variable. In their experiments, the authors find SSSMs to perform better than regular LDSs on a physiological data modeling task with multiple distinct underlying dynamical models. HMMs performed comparably well in terms of log-likelihood, indicating their ability to model nonlinear dynamics though the resulting model was less interpretable than the best SSSM\@. More recently, models for nonlinear time series modeling such as Gaussian Process Dynamical Models have been proposed~\cite{wang05}. However, the parameter learning algorithm is only locally optimal, and exact inference and simulation are very expensive, requiring MCMC over a long sequence of frames all at once. This necessitates the use of heuristics for both inference and learning. Another recent nonlinear dynamic model is~\cite{langford09}, which differs greatly from other methods in that it treats each component of the dynamic model learning problem separately using supervised learning algorithms, and proves consistency on the aggregate result under certain strong assumptions.
\section{Discussion}
The spectral learning algorithm blurs the line between latent variable models and PSRs. PSRs were developed with a focus on the problem of an agent planning actions in a partially observable environment. More generally, there are many scenarios in sequential data modeling where the underlying dynamical system has inputs. The inference task for a learned model is then to track the belief state while conditioning on observations and incorporating the inputs. The input-output HMM (IO-HMM)~\cite{bengio95} is a conditional probabilistic model which has these properties. A natural generalization of this work is to the task of learning RR-HMMs with inputs, or controlled PSRs. We recently carried out this generalization to controlled PSRs; details can be found in~\cite{boots09arxiv}.

The question of proving containment or equivalence of RR-HMMs with respect to PSRs is of theoretical interest. The observable representation of an RR-HMM is a Transformed PSR (TPSR)~\cite{rosencrantz04}, so every RR-HMM is a PSR; it remains to be seen whether every PSR corresponds to some RR-HMM (possibly with an infinite number of discrete hidden states) as well. The idea that ``difficult'' PSRs should somehow correspond to RR-HMMs with very large or infinite state space is intuitively appealing but not straightforward to prove. Another interesting direction would be to bound the performance of the learning algorithm when the underlying model is only approximately a reduced-rank HMM, much as the HKZ algorithm includes bounds when the underlying model is approximately an HMM~\cite{zhang09}. This would be useful since in practice it is more realistic to expect any underlying system to not comply with the exact model assumptions.

The positive realization problem, i.e.\@ obtaining stochastic transition and observation matrices from the RR-HMM observable representation, is also significant, though the observable representation allows us to carry out all possible HMM operations. HKZ describes a method based on~\cite{mossel06} which, however, is highly erratic in practice. In the RR-HMM case, we have the additional challenge of firstly computing the minimal $m$ for which a positive realization exists, and since the algorithm learns PSRs there is no guarantee that a particular set of learned parameters conforms exactly to any RR-HMM. On the applications side, it would be interesting to compare RR-HMMs with other dynamical models on classification tasks, as well as on learning models of difficult video modeling and graphics problems for simulation purposes. More elaborate choices of features may be useful in such applications, as would be the usage of high-dimensional or infinite-dimensional features via Reducing Kernel Hilbert Spaces (RKHS).
\section*{Acknowledgements}
We acknowledge helpful conversations with Sham Kakade regarding the HKZ spectral learning algorithm~\cite{zhang09}. Julian Ramos assisted with the gathering of robot vision data used in our experiments. SMS was supported by the NSF under grant number 0000164, the USAF under grant number FA8650-05-C-7264, the USDA under grant number 4400161514, and a project with MobileFusion/TTC\@. BEB was supported by the NSF under grant number EEEC-0540865. GJG was supported by DARPA under grant number HR0011-07-10026, the Computer Science Study Panel program, and by DARPA/ARO under MURI grant number W911NF-08-1-0301. BEB and GJG were both supported by ONR MURI grant number N00014-09-1-1052.
\bibliography{RRHMM}
\bibliographystyle{unsrt}
\section{Appendix I: Proofs}
The proof of Theorem~\ref{thm_HKZ_6} relies on Lemmas~\ref{lem_HKZ_8} and~\ref{lem_HKZ_12}.
We start off with some preliminary results and build up to proving the main theorem and its lemmas below.

\underline{A remark on norms}: The notation $\pnorm{X}{p}$ for \textit{matrices} $X \in \Rbbmat{m}{n}$ denotes the \textit{operator norm} $\max \frac{\pnorm{Xv}{p}}{\pnorm{v}{p}}$ for vector $v \neq 0$. Specifically, $\pnorm{X}{2}$ denotes \textit{$L_2$ matrix norm} (also known as \textit{spectral norm}), which corresponds to the \textit{largest singular value} $\sigma_1(X)$. \textit{Frobenius norm} is denoted by $\pnorm{X}{F} = \left(\sum_{i=1}^m \sum_{j=1}^n X_{ij}^2\right)^{1/2} $. The notation $\pnorm{X}{1}$ for matrices denotes the $L_1$ matrix norm which corresponds to \textit{maximum absolute column sum} $\max_c \sum_{i=1}^m \abs{X_{ic}}$. The definition of $\pnorm{x}{p}$ for \textit{vectors} $x\in\Rbbvec{n}$ is the standard distance measure $\left(\sum_{i=1}^n x_i^p\right)^{1/p}$.
\subsection{Preliminaries}
\label{RRHMM-sec-prelims} The following conditions are assumed by the main theorems and algorithms.

\begin{condition}\label{con_HKZ_1} [Modification of HKZ Condition 1] $\vec{\pi} > 0$ element-wise, $T$ has rank $k$ (i.e.\@ $R$ and $S$ both have rank $k$) and $O$ has rank at least $k$.
\end{condition}

The following two conditions on $R$ can always be satisfied by scaling down entries in $R$ and scaling up $S$ accordingly. However we want entries in $R$ to be as large as possible under the two conditions below, so that $\sigma_k\UOR$ is large and $1/\sigma_k\UOR$ is small to make the error bound as tight as possible (Theorem~\ref{thm_HKZ_6}). Hence we pay for scaling down $R$ by loosening the error bound we obtain for a given number of training samples.
\begin{condition}\label{con_R_L1} $\pnorm{R}{1} \leq 1$.
\end{condition}
\begin{condition}\label{con_R_unif} For some column $1 \leq c \leq k$ of $R$, it is the case that $\pnorm{R[\cdot,c]}{2} \leq \sqrt{k/m}$.
\end{condition}
The above two conditions on $R$ ensure the bounds go through largely unchanged from HKZ aside from the improvement due to low rank $k$. The first condition can be satisfied in a variety of ways without loss of generality, e.g.\@ by choosing the columns of $R$ to be any $k$ independent columns of $T$, and $S$ to be the coefficients needed to reconstruct $T$ from $R$. Intuitively, the first condition implies that $R$ does not overly magnify the magnitude of vectors it multiplies with. The second one implies a certain degree of uniformity in at least one of the columns of $R$. For example, the uniform distribution in a column of $R$ would satisfy the constraint, whereas a column of the identity matrix would not. This does not imply that $T$ must have a similarly near-uniform column. We can form $R$ from the uniform distribution along with some independent columns of T.

The observable representation depends on a matrix $U \in \Rbbmat{n}{k}$ that obeys the following condition:
\begin{condition}\label{con_HKZ_2} [Modification of HKZ Condition 2] $U\tps  O R$ is invertible.
\end{condition}
This is analogous to the HKZ invertibility condition on $U\tps  O$, since $OR$ is the matrix that yields observation probabilities from a \textit{low-dimensional} state vector. Hence, $U$ defines a $k$-dimensional subspace that preserves the \textit{low-dimensional} state dynamics regardless of the number of states $m$.

\begin{condition}\label{con_SpiO} Assume that $S\diag(\pivec)O\tps$ has full row rank (i.e.\@ $k$).
\end{condition}
This condition amounts to ensuring that that the ranges $S$ and $O$, which are both at least rank $k$, overlap enough to preserve the dynamics when mapping down to the low-dimensional state.
As in HKZ, the left singular vectors of $\Ptwoone$ give us a valid $U$ matrix.
%
\begin{lemma}\label{lem_HKZ_2} [Modification of HKZ Lemma 2] Assume Conditions~\ref{con_HKZ_1} and~\ref{con_SpiO}. Then, $\rank(\Ptwoone) = k$. Also, if $U$ is the matrix of left singular vectors of $\Ptwoone$ corresponding to non-zero singular values, then $\range(U) = \range(OR)$, so $U \in \mathbb{R}^{n\times k}$ obeys Condition~\ref{con_HKZ_2}.
\end{lemma}
\begin{proof} From its definition, we can show $\Ptwoone$ can be written as a low-rank product of RR-HMM parameters:
\begin{align}\label{eq:Ptwoone}
[\Ptwoone]_{i,j} &= \Pr[x_2 = i, x_1 = j] \nonumber\\
&= \sum_{a=1}^m \sum_{b=1}^m \Pr[x_2 = i, x_1 = j, h_2 = a, h_1 = b] \quad \mbox{(marginalizing hidden states $h$)}\nonumber\\
&= \sum_{a=1}^m \sum_{b=1}^m \Pr[x_2 = i | h_2 = a]\Pr[h_2 = a | h_1 = b] \Pr[ x_1 = j | h_1 = b]\Pr[h_1 = b] \nonumber\\
&= \sum_{a=1}^m \sum_{b=1}^m O_{ia} T_{ab} \pivec_b [O\tps]_{bj} \nonumber\\
\Rightarrow \Ptwoone &= OT \diag(\pivec)O\tps \nonumber\\
&= ORS \diag(\pivec)O\tps
\end{align}
Thus $\mathrm{range}(\Ptwoone) \subseteq \range(OR)$. This shows that $\rank(\Ptwoone) \leq \rank(OR)$.

By Condition~\ref{con_SpiO},  $S\diag(\pivec)O\tps$ has full row rank, thus  $S\diag(\vec{\pi})O\tps (S\diag(\vec{\pi})O\tps )^+ = I_{k\times k}$. Therefore,
\begin{align}
\label{eq:OR_P21} OR &= \Ptwoone(S\diag(\vec{\pi})O\tps )^+
\end{align}
which implies $\range(OR) \subseteq \range(\Ptwoone)$, which in turn implies  $\rank(OR) \leq \rank(\Ptwoone)$.

Together this proves that $\rank(\Ptwoone) = \rank(OR)$, which we can show to be $k$ as follows: Condition~\ref{con_HKZ_1} implies that $\rank\UOR = k$, and hence $\rank(OR) \geq k$. Since $OR \in \Rbbmat{m}{k}$, $\rank(OR) \leq k$. Therefore, $\rank(OR) = k$. Hence $\rank(\Ptwoone) = k$.

Since $\range(U) = \range(\Ptwoone)$ by definition of singular vectors, this implies $\range(U) = \range(OR)$. Therefore, $\UOR$ is invertible and hence $U$ obeys Condition~\ref{con_HKZ_2}.
\end{proof} \\

The following lemma shows that the observable representation $\{\binf,\bone,B_1,\ldots,B_n\}$ is linearly related to the true HMM parameters, and can compute the probability of any sequence of observations.
\begin{lemma}\label{lem_HKZ_3}[Modification of HKZ Lemma 3] (Observable HMM Representation). Assume Condition~\ref{con_HKZ_1} on the RR-HMM and Condition~\ref{con_HKZ_2} on the matrix $U \in \mathbb{R}^{n\times k}$. Then, the \textit{observable representation} of the RR-HMM (Definition 1 of the paper) has the following properties:
\begin{enumerate}
\item $\bone\quad =\quad \UOR\pi_l \quad =\quad \UO\pi$,
\item $\binft\quad =\quad 1_m\tps  R \UORi$,
\item For all $x = 1,\ldots,n:\ B_x \quad =\quad  \UOR W_x \UORi$
\item For any time $t$: $\Pr[x_{1:t}] \quad = \quad \binft  B_{x_t:1} \bone$
\end{enumerate}
\end{lemma}

\begin{proof}

\begin{enumerate}
\item We can write $\Pone$ as $O\pi$, since
\begin{align*}
[\Pone]_i &= \Pr[x_1 = i] \\
&= \sum_{a=1}^m \Pr[x_1 = i | h_1 =  a]\Pr[h_1 = a] \\
&= \sum_{a=1}^m O_{ia} \pivec_a
\end{align*}
Combined with the fact that $\bone = U\tps \Pone$ by definition, this proves the first claim.

\item Firstly note that $\Pone\tps = \onevect_m T \diag(\pivec) O\tps$, since
\begin{align*}
\Pone\tps &= \pivec\tps O\tps \\
&= \onevect_m \diag(\pivec) O\tps \\
&= \onevect_m T \diag(\pivec) O\tps \quad \mbox{(since $\onevect_m T = \onevect_m$)}
\end{align*}
This allows us to write $\Pone$ in the following form:
\begin{align*}
\Pone\tps &= \onevect_m T \diag(\pi) O\tps  \\
&= \onevect_m R S \diag(\pi) O\tps  \\
&= \onevect_m R \UORi \UOR S \diag(\pi) O\tps  \\
&= \onevect_m R \UORi U\tps  \Ptwoone \quad \mbox{(by equation~\eqref{eq:Ptwoone})}
\end{align*}
By equation~\eqref{eq:Ptwoone}, $U\tps \Ptwoone = \UOR S\diag(\pivec)O\tps$. Since $\UOR$ is invertible by Condition~\ref{con_HKZ_2}, and $S\diag(\pivec)O\tps$ has full row rank by Condition~\ref{con_SpiO}, we know that $(U\tps \Ptwoone)^+$ exists and
\begin{align}
\label{eq:up21} U\tps \Ptwoone (U\tps \Ptwoone)^+ &= I_{k\times k}
\end{align}
Therefore,
\begin{align*}
b\tps_\infty = P\tps_1 (U\tps  \Ptwoone)^+ = 1\tps_m R \UORi (U\tps  \Ptwoone) (U\tps  \Ptwoone)^+ = 1\tps_m R \UORi
\end{align*}
hence proving the second claim.
\item The third claim can be proven by first expressing $\Pthreexone$  as a product of RR-HMM parameters:
\begin{align*}
[\Pthreexone]_{ij} &= \Pr[x_3 = i, x_2 = x, x_1 = j] \\
 &= \sum_{a=1}^m \sum_{b=1}^m \sum_{c=1}^m \Pr[x_3 = i, x_2 = x, x_1 = j, h_3 = a, h_2 = b, h_1 = c] \\
&=  \sum_{a=1}^m \sum_{b=1}^m \sum_{c=1}^m \Pr[x_3 = i | h_3 = a] \Pr[h_3 = a | h_2 = b] \Pr[x_2 = x | h_2 = b] \\
&\ \ \ \ \ \ \ \ \ \ \ \ \ \ \ \ \ \ \ \ \ \ \ \  \Pr[h_2 = b | h_1 = c]  \Pr[h_1=c] \Pr[x_1 = j | h_1 = c] \\
&=  \sum_{a=1}^m \sum_{b=1}^m \sum_{c=1}^m O_{ia} [A_x]_{ab} T_{bc}  \pivec_c [O\tps]_{cj}\\
\Rightarrow \Pthreexone &=  OA_x T \diag(\pivec)O\tps
\end{align*}
This can be transformed as follows:
\begin{align*}
\Pthreexone &=  OA_x R S \diag(\pivec)O\tps \\
&=  OA_x R \UORi \UOR S \diag(\pivec)O\tps \\
&=  OA_x R \UORi U\tps (O T \diag(\pivec)O\tps) \\
&=  OA_x R \UORi U\tps \Ptwoone \quad \mbox{(by equation~\eqref{eq:Ptwoone})} \\
\end{align*}

and then plugging in this expression into the definition of $B_x$, we obtain the required result:
\begin{align*}
B_x &= (U\tps \Pthreexone)(U\tps \Ptwoone)^+ \quad \mbox{(by definition)} \\
&= \UO A_x R \UORi (U\tps \Ptwoone)(U\tps \Ptwoone)^+ \\
&= \UO A_x R \UORi \quad \mbox{(by equation~\eqref{eq:up21})}\\
&= \UOR \left(S \diag(O_{x,\cdot}) R\right) \UORi \\
&= \UOR W_x \UORi \\
\end{align*}

\item Using the above three results, the fourth claim follows from equation 1 in Section 2 in the paper:
\begin{align*}
& \Pr[x_1,\ldots,x_t] \\
&= \onevect_m R W_{x_t} \ldots W_{x_1} \pivec_l \\
&= \onevect_m R\UORi \UOR W_{x_t} \UORi \UOR W_{x_{t-1}} \UORi \ldots \\
&\ \ \ \ \ \ \ldots \UOR W_{x_1} \UORi \UOR \pivec_l \\
&= \binft B_{x_t} \ldots B_{x_1} \bone \\
&= \binft B_{x_{t:1}}\bone
\end{align*}
\end{enumerate}
\end{proof} \\
In addition to $\bone$ above, we define normalized conditional `internal states' $\bvec_t$ that help us compute conditional probabilities. These internal states are not probabilities. In contrast to HKZ where these internal states are $m$-dimensional vectors, in our case the internal states are $k$-dimensional i.e.\@ they correspond to the rank of the HMM. As shown above in Lemma~\ref{lem_HKZ_3},
\[\bone = \UOR\pivec_l = \UO \pivec\]
 In addition for any $t\geq 1$, given observations $x_{1:t-1}$ with non-zero probability, the internal state is defined as:
\begin{align}
\label{eq:state_update} \bvec_t &= \bvec_t(x_{1:t-1}) = \frac{B_{x_{t-1:1}}\bone}{\binft B_{x_{t-1}:1}\bone}
\end{align}
For $t=1$ the formula is still consistent since $\binft \bone = \onevect_m R \UORi\UOR\pivec_l = \onevect_m R \pivec_l = \onevect_m\pivec = 1$.

Recall that HMM and RR-HMM parameters can be used to calculate joint probabilities as follows:
\begin{align}
\label{eq_appendix:jointprobcalc}
\Pr[x_1,...,x_t] &= \onevect_m A_{x_t} A_{x_{t-1}} \cdots  A_{x_{1}} \pivec \nonumber \\
&= \onevect_m RS \diag(O_{ x_t,\cdot}) RS \diag(O_{ x_{t-1},\cdot})R \cdots S \diag(O_{ x_{1},\cdot}) \pivec\nonumber\\
&= \onevect_m R \left(S \diag(O_{ x_t,\cdot}) R\right) \left(S \diag(O_{ x_{t-1},\cdot})R\right) \cdots \left(S \diag(O_{ x_{1},\cdot})R\right) S \pivec\nonumber\\
&= \onevect_m  R W_{x_t} \ldots W_{x_1} \pivec_l\quad\mbox{(by definition of $W_x,\pivec_l$)}
\end{align}
The following Lemma shows that the conditional internal states are linearly related, and also shows how we can use them to compute conditional probabilities.
\begin{lemma}\label{lem_HKZ_4}[Modification of HKZ Lemma 4] (Conditional Internal States) Assume the conditions of Lemma~\ref{lem_HKZ_3} hold, i.e. Conditions~\ref{con_HKZ_1} and~\ref{con_HKZ_2} hold. Then, for any time $t$:
\begin{enumerate}
\item (Recursive update) If $\Pr[x_1,\ldots,x_t] > 0$, then
\begin{align*}
\bvec_{t+1} &= \frac{B_{x_t} \bvec_t}{\binft B_{x_t} \bvec_t}
\end{align*}
\item (Relation to hidden states)
\begin{align*}
\bvec_t &= \UOR l_t(x_{1:t-1}) = \UO h_t(x_{1:t-1})
\end{align*}
where $[\hvec_t(x_{1:t-1})]_i = \Pr[h_t = i | x_{1:t-1}]$ is defined as the conditional probability of the hidden state at time $t$ given observations $x_{1:t-1}$, and $\lvec_t(x_{1:t-1})$ is its low-dimensional projection such that $\hvec_t(x_{1:t-1}) = R\lvec_t(x_{1:t-1})$.
\item (Conditional observation probabilities)
\begin{align*}
\Pr[x_t | x_{1:t-1}] &= \binft B_{x_t} \bvec_t
\end{align*}
\end{enumerate}
\end{lemma}
\begin{proof}
The first proof is direct, the second follows by induction.
\begin{enumerate}
\item The $t=2$ case $\bvec_{2} = \frac{B_{x_1}\bone}{\binft B_{x_1}\bone}$ is true by definition (equation~\ref{eq:state_update}). For $t \geq 3$, again by definition of $b_{t+1}$ we have
\begin{align*}
\bvec_{t+1} &=\frac{B_{x_{t:1}}\bone}{\binft B_{x_{t:1}}\bone} \\
&= \frac{B_{x_t}B_{x_{t-1:1}}\bone}{\frac{\binft B_{x_{t-1:1}}\bone}{\binft B_{x_{t-1:1}}\bone}\binft B_{x_t}B_{x_{t-1:1}}\bone} \\
&= \frac{B_{x_t}\bvec_t}{\binft B_{x_t} \frac{B_{x_{t-1:1}}\bone}{\binft B_{x_{t-1:1}\bone}}} \quad\mbox{(by equation~\eqref{eq:state_update})} \\
&= \frac{B_{x_t}\bvec_t}{\binft B_{x_t}\bvec_t} \quad\mbox{(by equation~\eqref{eq:state_update})}
\end{align*}
\item[2,3.] The base case for claim 2 holds by Lemma~\ref{lem_HKZ_3}, since $\hvec_1 = \pivec$, $\lvec_1 = R\pivec$ and $\bone = \UOR\pivec$. For claim 3, the base case holds since $\binft B_{x_1} \bone = \onevect_m R W_{x_1} \pivec_l$ by Lemma~\ref{lem_HKZ_3}, which equals $\Pr[x_1]$ by equation~\eqref{eq_appendix:jointprobcalc}. The inductive step is:
\begin{align*}
\bvec_{t+1} &= \frac{B_{x_t}\bvec_t}{\binft B_{x_t}\bvec_t} \quad \mbox{(by claim 1 above)}\\
&= \frac{B_{x_t}\UOR\lvec_t}{\Pr[x_t|x_{1:t-1}]} \quad \mbox{(by inductive hypothesis)}\\
&= \frac{\UOR W_{x_t} \lvec_t}{\Pr[x_t|x_{1:t-1}]} \quad \mbox{(by Lemma~\ref{lem_HKZ_3})}\\
&= \frac{\UO A_{x_t} \hvec_t}{\Pr[x_t|x_{1:t-1}]}\quad\mbox{($\because RW_{x_t}\lvec_t = RS\diag(O_{x_t,\cdot})R\lvec_t = A_{x_t}\lvec_t$)}
\end{align*}
Now by definition of $A_{x_t}\hvec_t$,
\begin{align*}
\bvec_{t+1}&= \UO\frac{ \Pr[h_{t+1} = \cdot, x_t|x_{1:t-1}]}{\Pr[x_t|x_{1:t-1}]} \\
&= \UO\frac{ \Pr[h_{t+1} = \cdot|x_{1:t}]\Pr[x_t|x_{1:t-1}]}{\Pr[x_t|x_{1:t-1}]} \\
&= \UO\hvec_{t+1}(x_{1:t}) \\
&= \UOR\lvec_{t+1}(x_{1:t})
\end{align*}
This proves claim 2, using which we can complete the proof for claim 3:
\begin{align*}
\binft B_{x_{t+1}}\bvec_{t+1} &= \onevect_m R \UORi \UOR W_{x_t} \UORi \bvec_{t+1} \quad \mbox{(by Lemma~\ref{lem_HKZ_3})}\\
&= \onevect_m R W_{x_t} \UORi \UOR \lvec_{t+1} \quad \mbox{(by claim 2 above)} \\
&= \onevect_m R W_{x_t} \lvec_{t+1} \\
&= \onevect_m R S \diag(O_{x_t,\cdot}) R\lvec_{t+1} \quad \mbox{(by definition of $W_{x_t}$)} \\
&= \onevect_m T\diag(O_{x_t,\cdot}) \hvec_{t+1} \\
&= \onevect_m A_{x_t} \hvec_{t+1}
\end{align*}
Again by definition of $A_{x_t} \hvec_{t+1}$,
\begin{align*}
\binft B_{x_{t+1}}\bvec_{t+1} &= \sum_{a=1}^m \sum_{b=1}^m \Pr[x_{t+1}|h_{t+1} = a]\Pr[h_{t+1} = a | h_t = b] \Pr[h_t = b | x_{1:t}] \\
&= \sum_{a=1}^m \sum_{b=1}^m \Pr[x_{t+1},h_{t+1} = a, h_t = b | x_{1:t}] \\
&= \Pr[x_{t+1} | x_{1:t}]
\end{align*}
\end{enumerate}
\end{proof}

\begin{remark} If $U$ is the matrix of left singular vectors of $\Ptwoone$ corresponding to non-zero singular values, then $U$ is the observable-representation analogue of the observation probability matrix $O$ in the sense that, given a conditional state $\bvec_t$, $\Pr[x_t = i | x_{1:t-1}] = [U\bvec_t]_i$ in the same way as $\Pr[x_t = i | x_{1:t-1}] = [O\hvec_t]_i$ for a conditional hidden state $\hvec_t$.
\end{remark}
\begin{proof}
Since $\range(U) = \range(OR)$ (Lemma 2), and $UU\tps$ is a projection operator to $\range(U)$, we have $UU\tps  OR = OR$, so $U \bvec_t = U\UOR l_t = OR l_t = O h_t$.
\end{proof}



\subsection{Matrix Perturbation Theory}
\label{RRHMM-sec-matrix} We take a diversion to matrix perturbation theory and state some standard theorems from Steward and Sun (1990)~\cite{stewart-sun:1990} and Wedin (1972)~\cite{wedin72} which we will use, and also prove a result from these theorems.
The following lemma bounds the $L_2$-norm difference between the pseudoinverse of a matrix and the pseudoinverse of its perturbation.
\begin{lemma}\label{lem_HKZ_23}(Theorem 3.8 of Stewart and Sun (1990)~\cite{stewart-sun:1990}) Let $A \in \mathbb{R}^{m\times n}$, with $m \geq n$, and let $\widetilde{A} = A + E$. Then,
\[ \pnorm{\widetilde{A}^+ - A^+}{2} \leq \frac{1+\sqrt{5}}{2}\cdot \max\left\{\pnorm{A^+}{2}^2,\pnorm{\widetilde{A}^+}{2}^2\right\}\pnorm{E}{2}\quad .\]
\end{lemma}
The following lemma bounds the absolute differences between the singular values of a matrix and its perturbation.
\begin{lemma}\label{lem_HKZ_20}(Theorem 4.11 of Stewart and Sun (1990)~\cite{stewart-sun:1990}). Let $A \in \mathbb{R}^{m\times n}$ with $m \geq n$, and let $\widetilde{A} = A + E$. If the singular values of $A$ and $\widetilde{A}$ are $(\sigma_1 \geq \ldots \geq \sigma_n)$ and $(\widetilde{\sigma}_1 \geq \ldots \geq \widetilde{\sigma}_n)$, respectively, then
\[ | \widetilde{\sigma}_i - \sigma_i | \leq \pnorm{E}{2} \ \ i = 1,\ldots,n \]
\end{lemma}
Before the next lemma we must define the notion of \textit{canonical angles} between two subspaces:
\begin{definition} \label{def:canonical_angles} (Adapted from definition 4.35 of Stewart (1998)~\cite{stewart:1998}) Let $X$ and $Y$ be matrices whose columns comprise orthonormal bases of two $p$-dimensional subspaces $\mathcal{X}$ and $\mathcal{Y}$ respectively. Let the singular values of $X\tps Y$ (where $X\tps$ denotes the conjugate transpose, or Hermitian, of matrix $X$) be $\gamma_1, \gamma_2, \ldots, \gamma_p$. Then the \emph{canonical angles} $\theta_i$ between $\mathcal{X}$ and $\mathcal{Y}$  are defined by
\[ \theta_i = \cos^{-1} \gamma_i, \quad i= 1,2,\ldots,p\]
The \emph{matrix of canonical angles} $\Theta$ is defined as
\[ \Theta(\mathcal{X},\mathcal{Y}) = \diag(\theta_1,\theta_2,\ldots,\theta_p)\]
\end{definition}
Note that $\forall i \ \ \gamma_i \in [0,1]$ in the above definition, since $\gamma_1$ (assuming it's the highest singular value) is no greater than $\sigma_1(X\tps)\sigma_1(Y) \leq 1\cdot 1 = 1$, and hence $\cos^{-1}\gamma_i$ is always well-defined.

For any matrix $A$, define $A_\perp$ to be the \emph{orthogonal complement} of the subspace spanned by the columns of $A$. For example, any subset of left singular vectors of a matrix comprise the orthogonal complement of the matrix composed of the remaining left singular vectors. The following lemma gives us a convenient way of calculating the sines of the canonical angles between two subspaces using orthogonal complements:
\begin{lemma} \label{lem:sin_canonical_angles}(Theorem 4.37 of Stewart (1998)~\cite{stewart:1998}) Let $X$ and $Y$ be $n\times p$ matrices, with $n > p$, whose columns comprise orthonormal bases of two $p$-dimensional subspaces $\mathcal{X}$ and $\mathcal{Y}$ respectively. Assume $X_\perp,Y_\perp \in \Rbbmat{n}{n-p}$ such that $[X \ X_\perp]$ and $[Y \ Y_\perp]$ are orthogonal matrices. The singular values of $Y\tps_\perp X$ are the sines of the canonical angles between $\mathcal{X}$ and $\mathcal{Y}$.
\end{lemma}
The following lemma bounds the $L_2$-norm difference between the sine of the canonical angle matrices of the range of a matrix and its perturbation.
\begin{lemma}\label{lem_HKZ_21}(\cite{wedin72},Theorem $4.4$ of Stewart and Sun (1990) ~\cite{stewart-sun:1990}). Let $A \in \mathbb{R}^{m\times n}$ with $m \geq n$, with the singular value decomposition ($U_1,U_2,U_3,\Sigma_1,\Sigma_2,V_1,V_2$):
\begin{align*}
\left[
\begin{array}{c}
U\tps_1 \\
U\tps_2 \\
U\tps_3 \\
\end{array}
\right] A \left[ \begin{array}{c c} V_1 & V_2 \end{array} \right] = \left[ \begin{array}{cc}
\Sigma_1 & 0 \\
0 & \Sigma_2 \\
0 & 0 \\
\end{array}
\right]
\end{align*}
Let $\widetilde{A} = A + E$, with analogous SVD ($\widetilde{U}_1,\widetilde{U}_2,\widetilde{U}_3,\widetilde{\Sigma}_1,\widetilde{\Sigma}_2,\widetilde{\Sigma}_3,\widetilde{V}_1,\widetilde{V}_2$). Let $\Phi$ be the matrix of canonical angles between $\range(U_1)$ and $\range(\widetilde{U}_1)$, and $\Theta$ be the matrix of canonical angles between $\range(V_1)$ and $\range(\widetilde{V}_1)$. If there exists $\delta > 0 ,\alpha \geq 0$ such that $\min \sigma(\widetilde{\Sigma}_1) \geq \alpha + \delta$ and $\max \sigma(\Sigma_2) \leq \alpha$, then
\[ \max\left\{\pnorm{\sin\Phi}{2}, \pnorm{\sin\Theta}{2}\right\} \leq \frac{\pnorm{E}{2}}{\delta} \]
\end{lemma}
The above two lemmas can be adapted to prove that Corollary 22 of HKZ holds for the low-rank case as well, assuming that the perturbation is bounded by a number less than $\sigma_k$. The following lemma shows that (1) the $k^{\mathrm{th}}$ singular value of a matrix and its perturbation are close to each other, and (2) that the  subspace spanned by the first $k$ singular vectors of a matrix is nearly orthogonal to the subspace spanned by the $(k+1)^{\mathrm{th}},\ldots,m^{\mathrm{th}}$ singular vectors of its perturbation, with the matrix product of their bases being bounded.
\begin{corollary}\label{cor_HKZ_22}[Modification of HKZ Corollary 22] Let $A \in  \Rbbmat{m}{n}$, with $m \geq n$, have rank $k < n$, and let $U \in \Rbbmat{m}{k}$ be the matrix of $k$ left singular vectors corresponding to the non-zero singular values $\sigma_1 \geq \ldots \geq \sigma_k \geq 0$ of $A$. Let $\widetilde{A} = A + E$. Let $\widetilde{U} \in \Rbbmat{m}{k}$ be the matrix of $k$ left singular vectors corresponding to the largest $k$  singular values $\widetilde{\sigma}_1 \geq \ldots \geq \widetilde{\sigma}_k$ of $\widetilde{A}$, and let $\widetilde{U}_\perp \in \Rbbmat{m}{(m - k)}$ be the remaining left singular vectors. Assume $\pnorm{E}{2} \leq \epsilon\sigma_k$ for some $\epsilon < 1$. Then:
\begin{enumerate}
\item $\widetilde{\sigma}_k \geq (1-\epsilon)\sigma_k$.
\item $\pnorm{\widetilde{U}\tps_\perp U}{2} \leq \pnorm{E}{2} / \widetilde{\sigma}_k$.
\end{enumerate}
\end{corollary}
\begin{proof}
\begin{enumerate}
\item From Lemma~\ref{lem_HKZ_20},
\begin{align*}
\abs{\widetilde{\sigma}_k - \sigma_k} &\leq \pnorm{E}{2} \\
\abs{\widetilde{\sigma}_k - \sigma_k} &\leq \epsilon\sigma_k \\
\widetilde{\sigma}_k - \sigma_k &\geq -\epsilon\sigma_k \\
\widetilde{\sigma}_k &\geq (1-\epsilon)\sigma_k
\end{align*}
which proves the first claim.
\item Recall that by Lemma~\ref{lem:sin_canonical_angles}, if $\Phi$ is a matrix of all canonical angles between $\range(\Ptwoone)$ and $\range(\Phtwoone)$, then $\sin \Phi$ contains all the singular values of $\widetilde{U}\tps_\perp U$ along its diagonal.

Also recall that the $L_2$ norm of a matrix is its top singular value. Then,
\begin{align*}
\pnorm{\sin \Phi}{2} &= \sigma_1(\sin \Phi) \quad\mbox{(by definition)}\\
&= \max\diag(\sin \Phi) \quad\mbox{(since $\sin \Phi$ is a diagonal matrix)} \\
&= \sigma_1(\Utilde\tps_\perp U)\quad\mbox{(by Lemma~\ref{lem:sin_canonical_angles})}\\
&= \pnorm{\Utilde\tps_\perp U}{2}\quad\mbox{(by definition)}
\end{align*}

Invoking Lemma~\ref{lem_HKZ_21} with the parameter values $\delta = \widetilde{\sigma}_k$ and $\alpha = 0$ yields $\pnorm{\sin \Phi}{2} \leq \pnorm{E}{2}/\widetilde{\sigma}_k$. Combining this with $\pnorm{\sin \Phi}{2} =  \pnorm{(\widetilde{U}\tps_\perp U)}{2}$ proves claim 2.
\end{enumerate}
\end{proof}
%
\subsection{Supporting Lemmas}
\label{RRHMM-sec-support} In this section we develop the main supporting lemmas that help us prove Theorem~\ref{thm_HKZ_6}

\subsubsection{Estimation Errors}
\label{RRHMM-sec-proofs-errs} We define $\eone$,$\etwoone$ and $\ethreexone$ as sampling errors for $\Pone$,$\Ptwoone$ and $\Pthreexone$ respectively:
\begin{subequations} \label{eq:epsdefs}
\begin{align}
\eone &= \pnorm{\Phone - \Pone}{F} \\
\etwoone &= \pnorm{\Phtwoone - \Ptwoone}{F} \\
\ethreexone &= \pnorm{\Phthreexone - \Pthreexone}{F} \quad \mbox{for $x = 1,\ldots,n$}
\end{align}
\end{subequations}
\begin{lemma}\label{lem_HKZ_8} [Modification of HKZ Lemma 8] If the algorithm independently samples $N$ observation triples from the HMM, then with probability at least $1-\eta$:
\begin{align*}
\eone &\leq \sqrt{\frac{1}{N}\ln\frac{3}{\eta}} + \sqrt{\frac{1}{N}} \\
\etwoone &\leq \sqrt{\frac{1}{N}\ln\frac{3}{\eta}} + \sqrt{\frac{1}{N}} \\
\max_{x} \ethreexone &\leq \sqrt{\frac{1}{N}\ln\frac{3}{\eta}} + \sqrt{\frac{1}{N}} \\
\sum_{x} \ethreexone &\leq \min_{k} \left( \sqrt{\frac{k}{N}\ln\frac{3}{\eta}} + \sqrt{\frac{k}{N}} + 2\epsilon(k)\right) + \sqrt{\frac{1}{N}\ln{\frac{3}{\eta}}} + \sqrt{\frac{1}{N}}
\end{align*}
\end{lemma}
Before proving this lemma, we need some definitions and a preliminary result. First, we restate \textit{McDiarmid's Inequality}~\cite{mcdiarmid89}:
\begin{theorem}
\label{thm_mcdiarmids_ineq} Let $Z_1,\ldots,Z_m$ be independent random variables all taking values in the set $\mathcal{Z}$. Let $c_i$ be some positive real numbers. Further, let $f:\mathcal{Z}^m \mapsto \mathbb{R}$ be a function of $Z_1,\ldots,Z_m$ that satisfies $\forall i$, $\forall z_1,\ldots,z_m,z'_i \in \mathcal{Z}$,
\[ \abs{f(z_1,\ldots,z_i,\ldots,z_m) - f(z_1,\ldots,z'_i,\ldots,z_m)} \leq c_i.\]
Then for all $\epsilon > 0$,
\[\Pr[f - \mathbb{E}[f] \geq \epsilon] \leq \exp\left(\frac{-2\epsilon^2}{\sum_{i=1}^m c_i^2}\right).\]
\end{theorem}

Assume $z$ is a discrete random variable that takes on values in ${1,\ldots,d}$. The goal is to estimate the vector $\qvec =  [\Pr(z =j)]_{j=1}^d$ from $N$ i.i.d.\@ samples $z_i\ (i = 1,\ldots,N)$. Let $e_j$ denote the $j^\mathrm{th}$ column of the $d\times d$ identity matrix. For $i = 1,\ldots,N$, suppose $\qvec_i$ is a column of the $d\times d$ identity matrix such that $\qvec_i(j) = e_{z_i}$. In other words, the $z_i^{\mathrm{th}}$ component of $\qvec_i$ is $1$ and the rest are $0$. Then the empirical estimate of $\qvec$ in terms of $\qvec_i$ is $\qhat = \sum_{i=1}^{N} \qvec_i/N$.

Each part of Lemma~\ref{lem_HKZ_8} corresponds to bounding, for some $\qvec$, the quantity
\[ \pnorm{\qhat - \qvec}{2}^2\quad . \]
We first state a result based on McDiarmid's inequality (Theorem~\ref{thm_mcdiarmids_ineq}):
\begin{proposition} \label{prop_HKZ_19} [Modification of HKZ Proposition 19]
For all $\epsilon > 0$ and $\qhat,\qvec$ and $N$ as defined above:
\[ \Pr\left( \pnorm{\qhat - \qvec}{2} \geq 1/\sqrt{N} + \epsilon\right) \leq e^{-N\epsilon^2} \]
\end{proposition}
\begin{proof}
Recall $\qhat = \sum_{i=1}^{N} \qvec_i/N$, and define $\phat = \sum_{i=1}^{N} \pvec_i/N$ where $\pvec_i = \qvec_i$ except for $i=k$, and $p_k$ is an arbitrary column of the appropriate-sized identity matrix. Then we have
\begin{align*}
\pnorm{\qhat - \qvec}{2} - \pnorm{\phat - \qvec}{2} &\leq \pnorm{\qhat - \phat}{2} \quad \mbox{(by triangle inequality)} \\
&= \pnorm{(\sum_i \qvec_i)/N - (\sum_i \pvec_i)/N}{2} \\
&= (1/N)\pnorm{\qvec_k - \pvec_k}{2}  \quad \mbox{(by definition of $\phat,\qhat$ and $L_2$-norm)}\\
&\leq (1/N)\sqrt{1^2 + 1^2} \\
&= \sqrt{2}/N
\end{align*}
This shows that $\pnorm{\qhat - \qvec}{2}$ is a function of random variables $\qvec_1,\ldots,\qvec_N$ such changing the $k^{\mathrm{th}}$ random variable $q_k$ for any $1 \leq k\leq N$ (resulting in $\pnorm{\phat - \qvec}{2}$) changes the value of the function by at most $c_k = \sqrt{2}/N$. Note that $\qvec$ is not a random variable but rather the variable we are trying to estimate. In this case, McDiarmid's inequality (Theorem~\ref{thm_mcdiarmids_ineq}) bounds the deviation $\pnorm{\qhat - \qvec}{2}$ from its expectation $\mathbb{E}\pnorm{\qhat - \qvec}{2}$ as:
\begin{align}
\label{eq:mcdiarmids}
\Pr(\pnorm{\qhat - \qvec}{2} \geq \mathbb{E}\pnorm{\qhat - \qvec}{2} + \epsilon) &\leq \exp{\frac{-2\epsilon^2}{\sum_{i=1}^N c_i^2} } \nonumber \\
&= \exp{\frac{-2\epsilon^2}{N\cdot 2 / N^2} }  \nonumber\\
&= e^{-N\epsilon^2}
\end{align}
We can bound the expected value using the following inequality:
\begin{align*}
\mathbb{E}\pnorm{\sum_{i=1}^N \qvec_i - N\qvec}{2} &= \mathbb{E}\left(\pnorm{\sum_{i=1}^N \qvec_i - N\qvec}{2}^2\right)^{1/2} \\
&\leq \left(\mathbb{E}\pnorm{\sum_{i=1}^N \qvec_i - N\qvec}{2}^2\right)^{1/2} \quad\mbox{(by concavity of square root, and Jensens inequality)} \\
&= \left(\sum_{i=1}^N \mathbb{E}\pnorm{ \qvec_i - \qvec}{2}^2\right)^{1/2} \\
&= \left(\sum_{i=1}^N \mathbb{E}(\qvec_i - \qvec)\tps(\qvec_i - \qvec)\right)^{1/2}
\end{align*}
Multiplying out and using linearity of expectation and properties of $\qvec_i$ (namely, that $\qvec_i\tps\qvec_i = 1$, $\mathbb{E}(\qvec_i) = \qvec$ and $\qvec$ is constant), we get:
\begin{align*}
\mathbb{E}\pnorm{\sum_{i=1}^N \qvec_i - N\qvec}{2} &\leq \left(\sum_{i=1}^N \mathbb{E}(1 - 2\qvec_i\tps\qvec +\pnorm{\qvec}{2}^2)\right)^{1/2}\quad\mbox{(since $\qvec_i\tps\qvec_i = 1$)}\\
&= \left(\sum_{i=1}^N \mathbb{E}(1) - 2\sum_{i=1}^N \mathbb{E}(\qvec_i\tps\qvec) +\sum_{i=1}^N \mathbb{E}\pnorm{\qvec}{2}^2\right)^{1/2} \\
&= \left(N - 2N\pnorm{\qvec}{2}^2 + N\pnorm{\qvec}{2}^2\right)^{1/2} \\
&= \sqrt{N(1 - \pnorm{\qvec}{2}^2)}
\end{align*}
This implies an upper bound on the expected value:
\begin{align*}
\mathbb{E}\pnorm{\qhat-\qvec}{2}^2 &= (1/N^2)\mathbb{E}\pnorm{\sum_{i=1}^N\qvec_i-N\qvec}{2}^2 \\
&\leq (1/N^2)\cdot N(1 - \pnorm{\qvec}{2}^2) \\
\Rightarrow \mathbb{E}\pnorm{\qhat-\qvec}{2} &\leq (1/\sqrt{N})\sqrt{(1 - \pnorm{\qvec}{2}^2)} \\
&\leq (1/\sqrt{N})
\end{align*}
Using this upper bound in McDiarmids inequality (equation~\eqref{eq:mcdiarmids}), we get a looser version of the bound that proves the proposition:
\begin{align*}
\Pr(\pnorm{\qhat - \qvec}{2} \geq 1/\sqrt{N} + \epsilon) &\leq e^{-N\epsilon^2}
\end{align*}
\end{proof}

We are now ready to prove Lemma~\ref{lem_HKZ_8}.

\begin{proof}[Lemma~\ref{lem_HKZ_8}]
We will treat $\widehat{P}_1$,$\widehat{P}_{2,1}$ and $\widehat{P}_{3,x,1}$ as vectors, and use McDiarmid's inequality to bound the error in estimating a distribution over a simplex based on indicator vector samples, using Proposition~\ref{prop_HKZ_19}. We know that
\[\Pr\left(\pnorm{\qhat - \qvec}{2} \geq  \sqrt{1/N} + \epsilon\right) \leq e^{-N\epsilon^2}\quad . \]
Now let $\eta = e^{-N\epsilon^2}$. This implies
\begin{align*}
\ln \eta &= -N\epsilon^2 \\
\ln (1/\eta) &= N\epsilon^2 \\
\epsilon &= \sqrt{\ln(1/\eta)/N}
\end{align*}
Hence,
\[\Pr\left(\pnorm{\qhat - \qvec}{2} \geq  \sqrt{1/N} + \sqrt{\ln(1/\eta)/N}\right) \leq \eta \]
Therefore, with probability at least $1-\eta$,
\begin{align}
\label{eq:propn_result} \pnorm{\qhat - \qvec}{2} &\leq 1/\sqrt{N} + \sqrt{\ln(1/\eta)/N}
\end{align}
Now, in place of $\qvec$ in equation~\eqref{eq:propn_result}, we substitute the stochastic vector $\Pone$ to prove the first claim, the vectorized version of the stochastic matrix $\Ptwoone$ to prove the second claim, and the vectorized version of the stochastic \textit{tensor} $\Pthreetwoone \in \mathbb{R}^{n\times n\times n}$ obtained by stacking $\Pthreexone$ matrices over all $x$, to prove the third claim. The matrices $\Phthreexone$ are stacked accordingly to obtain the estimated tensor $\Phthreetwoone$. We get the following:
\begin{align*}
\eone &\leq 1/\sqrt{N} + \sqrt{\ln(1/\eta)/N} \quad \mbox{(hence proving the first claim)}\\
\etwoone &\leq 1/\sqrt{N} + \sqrt{\ln(1/\eta)/N}  \quad \mbox{(hence proving the second claim)}\\
\max_x \ethreexone &\leq \sqrt{\sum_x \ethreexone^2} \\
&= \sqrt{\sum_x \pnorm{\Pthreexone - \Phthreexone}{2}^2}  \\
&= \sqrt{\sum_x \sum_i \sum_j ([\Pthreexone]_{i,j} - [\Phthreexone]_{i,j})^2}  \\
&= \sqrt{\pnorm{\Pthreetwoone -  \Phthreetwoone}{2}^2}  \\
&= \pnorm{\Pthreetwoone -  \Phthreetwoone}{2}  \\
&\leq \sqrt{1/N} + \sqrt{\ln(1/\eta)/N} \quad \mbox{(hence proving the third claim)}
\end{align*}
Note the following useful inequality from the above proof:
\begin{align}
\label{eq:sqrt_sum_e3} \sqrt{\sum_x \ethreexone^2} &\leq \sqrt{1/N} + \sqrt{\ln(1/\eta)/N}
\end{align}

It remains to prove the fourth claim, regarding $\sum_x \ethreexone$. First we get a bound that depends on $n$ as follows:
\begin{align*}
\sum_x \ethreexone &= \sum_x \abs{\ethreexone} \quad \mbox{($\because \forall x, \ethreexone \geq 0$)}\\
&\leq \sqrt{n}\sqrt{\sum_x \ethreexone^2} \quad \mbox{($\because \forall \xvec \in \Rbbvec{a}$, $\pnorm{\xvec}{1} \leq \sqrt{a}\pnorm{\xvec}{2}$)}\\
&\leq \sqrt{n/N} + \sqrt{\frac{n}{N}\ln \frac{1}{\eta}}
\end{align*}
We aren't going to use the above bound. Instead, if $n$ is large and $N$ small, this bound can be improved by removing direct dependence on $n$. Let $\epsilon(k)$ be the sum of smallest $n-k$ probabilities of the second observation $x_2$. Let $S_k$ be the set of these $n-k$ such observations $x$, for any $k$. Therefore,
\[ \epsilon(k) = \sum_{x \in S_k} \Pr[x_2 = x] =  \sum_{x \in S_k} \sum_{i,j} [\Pthreexone]_{ij} \]
Now, first note that we can bound $\sum_{x\notin S_k}\ethreexone$ as follows:
\begin{align*}
\sum_{x\notin S_k} \ethreexone &\leq \sum_{x\notin S_k} \left|\ethreexone\right| \\
&\leq \sqrt{k}\sqrt{\sum_{x\notin S_k} \ethreexone^2} \quad \mbox{($\because \forall \xvec \in \Rbbvec{a}$, $\pnorm{\xvec}{1} \leq \sqrt{a}\pnorm{\xvec}{2}$)}
\end{align*}
By combining with equation~\ref{eq:sqrt_sum_e3}, we get
\begin{align}
\label{eq:sqrt_sum_e3notSk} \sum_{x\notin S_k} \ethreexone &\leq \sqrt{k/N} + \sqrt{k\ln(1/\eta)/N}
\end{align}
To bound $\sum_{x\in S_k} \ethreexone$, we first apply equation~\eqref{eq:propn_result} again. Consider the vector $\qvec$ of length $kn^2 + 1$ whose first $kn^2$ entries comprise the elements of $\Pthreexone$ for all $x\notin S_k$, and whose last entry is the cumulative sum of elements of $\Pthreexone$ for all $x\in S_k$. Define $\qhat$ accordingly with $\Phthreexone$ instead of $\Pthreexone$. Now equation~\eqref{eq:propn_result} directly gives us with probability at least $1-\eta$:
\begin{align*}
\left[\sum_{x \notin S_k} \sum_{i,j}([\Phthreexone]_{i,j} - [\Pthreexone]_{i,j})^2 + \left|\sum_{x \in S_k}\sum_{i,j} ([\Phthreexone]_{ij} - \sum_{x \in S_k}\sum_{i,j}[\Pthreexone]_{ij})\right|^2\right]^\frac{1}{2} \leq \sqrt{1/N} + &\sqrt{\ln(1/\eta)/N} \\
\sum_{x \notin S_k} \pnorm{\Phthreexone - \Pthreexone}{F}^2 + \left|\sum_{x \in S_k}\sum_{i,j} ([\Phthreexone]_{ij} - [\Pthreexone]_{ij})\right|^2 \leq \left(\sqrt{1/N} + \sqrt{\ln(1/\eta)/N}\right)^2& \\
\end{align*}
Since the first term above is positive, we get
\begin{align}
\label{eq:absdiff_e3Sk} \left|\sum_{x \in S_k}\sum_{i,j}( [\Phthreexone]_{ij} - [\Pthreexone]_{ij})\right| &\leq \sqrt{1/N} + \sqrt{\ln(1/\eta)/N}
\end{align}
Now, by definition of $S_k$:,
\begin{align*}
\sum_{x \in S_k}\ethreexone &= \sum_{x \in S_k} \pnorm{\Phthreexone - \Pthreexone}{F} \\
&\leq \sum_{x \in S_k} \sum_{i,j} \left| [\Phthreexone]_{ij} - [\Pthreexone]_{ij}\right| \quad \mbox{($\because \forall \xvec $, $\pnorm{\xvec}{2} \leq \pnorm{\xvec}{1}$)}\\
&= \sum_{x \in S_k} \sum_{i,j} \max \left(0, [\Phthreexone]_{ij} - [\Pthreexone]_{ij}\right) \\
&\quad -  \sum_{x \in S_k} \sum_{i,j} \min \left(0, [\Phthreexone]_{ij} - [\Pthreexone]_{ij}\right) \quad\mbox{($\because \forall \xvec,|\xvec| = \left[\max(0,\xvec) - \min(0,\xvec)\right]$)}\\
&\leq \sum_{x \in S_k} \sum_{i,j} \max \left(0, [\Phthreexone]_{ij} - [\Pthreexone]_{ij}\right) + \sum_{x\in S_k}\sum_{i,j} [\Pthreexone]_{ij} \\
&\quad +  \sum_{x \in S_k} \sum_{i,j} \min \left(0, [\Phthreexone]_{ij} - [\Pthreexone]_{ij}\right) + \sum_{x\in S_k}\sum_{i,j} [\Pthreexone]_{ij} \\
&= \sum_{x \in S_k} \sum_{i,j} \max \left(0, [\Phthreexone]_{ij} - [\Pthreexone]_{ij}\right) + \epsilon(k) \\
&\quad +  \sum_{x \in S_k} \sum_{i,j} \min \left(0, [\Phthreexone]_{ij} - [\Pthreexone]_{ij}\right) + \epsilon(k) \quad\mbox{(by definition of $\epsilon(k)$)}\\
&\leq \left|\sum_{x \in S_k} \sum_{i,j} \left([\Phthreexone]_{ij} - [\Pthreexone]_{ij}\right)\right| + 2\epsilon(k)
\end{align*}
Plugging in equation~\eqref{eq:absdiff_e3Sk}, we get a bound on $\sum_{x \in S_k}\ethreexone$:
\begin{align*}
\sum_{x \in S_k}\ethreexone &\leq \sqrt{1/N} + \sqrt{\ln(1/\eta)/N} + 2\epsilon(k)
\end{align*}
Combining with equation~\eqref{eq:sqrt_sum_e3notSk} and noting that $k$ is arbitrary, we get the desired bound:
\[\sum_x \ethreexone \leq \min_{k}[\sqrt{k\ln(1/\eta)/N} + \sqrt{k/N} + \sqrt{\ln(1/\eta)/N} + \sqrt{1/N} + 2\epsilon(k)]\]

Note that, to get the term $\ln(3/\eta)$ instead of $\ln(1/\eta)$ as in the fourth claim, we simply use $\eta/3$ instead of $\eta$. This bound on $\sum_x \ethreexone$ will be small if the number of frequently occurring observations is small, even if $n$ itself is large.
\end{proof}

The next lemma uses the perturbation bound in Corollary~\ref{cor_HKZ_22} to bound the effect of sampling error on the estimate $\widehat{U}$, and on the conditioning of $\UhOR$.
\begin{lemma}\label{lem_HKZ_9}[Modification of HKZ Lemma 9] Suppose $\epsilon_{2,1} \leq \varepsilon\cdot\sPtwoone $ for some $\varepsilon < 1/2$. Let $\varepsilon_0 = \epsilon^2_{2,1}/((1-\varepsilon)\sPtwoone )^2$. Define $U,\Uhat \in \Rbbmat{m}{k}$ as the matrices of the first $k$ left singular vectors of $\Ptwoone,\Phtwoone$ respectively. Let $\theta_1,\ldots,\theta_k$ be the canonical angles between $\spann(U)$ and $\spann(\Uhat)$. Then:
\begin{enumerate}
\item $\varepsilon_0 < 1$
\item $\sigma_k(\widehat{U}\tps \widehat{P}_{2,1}) \geq (1-\varepsilon)\sPtwoone $
\item $\sigma_k(\widehat{U}\tps  \Ptwoone) \geq \sqrt{1-\varepsilon_0}\sPtwoone $
\item $\sigma_k(\widehat{U}\tps  OR) \geq \sqrt{1-\varepsilon_0}\sOR $
\end{enumerate}
\end{lemma}
\begin{proof}
First some additional definitions and notation. Define $\Uhat_\perp$ to be the remaining $n-k$ left singular vectors of $\Phtwoone$ corresponding to the lower $n-k$ singular values, and correspondingly $U_\perp$ for $\Ptwoone$. Suppose $U \Sigma V\tps = \Ptwoone$ is the thin SVD of $\Ptwoone$. Finally, we use the notation $\nuvec_i\{A\}\in \Rbbvec{q}$ to denote the $i^{\mathrm{th}}$ \textit{right singular vector} of a matrix $A \in \Rbbmat{p}{q}$. Recall that $\sigma_i(A) = \pnorm{A\nuvec_i\{A\}}{2}$ by definition.
\\
\\
\underline{First claim:} $\varepsilon_0 < 1$ follows from the assumptions:
\begin{align*}
\varepsilon_0 &= \frac{\etwoone^2}{((1-\varepsilon)\sPtwoone )^2} \\
&\leq \frac{\varepsilon^2 \sPtwoone ^2}{(1-\varepsilon)^2\sPtwoone ^2} \\
&= \frac{\varepsilon^2}{(1-\varepsilon)^2} \\
&< 1 \quad \mbox{(since $\varepsilon < 1/2$)}
\end{align*}

\underline{Second claim:} By Corollary~\ref{cor_HKZ_22}, $\sigma_k(\Phtwoone) \geq (1-\varepsilon)\sPtwoone $. The second claim follows from noting that $\sigma_k(\Uhat\tps \Phtwoone) =\sigma_k(\Phtwoone)$.

\underline{Third and fourth claims:} First consider the $k^{\mathrm{th}}$ singular value of $\Uhat\tps U$. For any vector $x\in \Rbbvec{k}$:
\begin{align*}
\frac{\pnorm{\Uhat\tps U x}{2}}{\pnorm{x}{2}} & \geq \min_y \frac{\pnorm{\Uhat\tps U y}{2}}{\pnorm{y}{2}} \\
&= \sigma_k(\Uhat\tps U) \quad\mbox{(by definition of smallest singular value)} \\
&= \cos(\theta_k) \quad\mbox{(by Definition~\ref{def:canonical_angles})} \\
&= \sqrt{1 - \sin^2(\theta_k)} \\
&= \sqrt{1 - \sigma_k(\Uhat_\perp\tps U)^2} \quad\mbox{(by Lemma~\ref{lem:sin_canonical_angles})}\\
&\geq \sqrt{1 - \sigma_1(\Uhat_\perp\tps U)^2}\\
&= \sqrt{1 - \pnorm{\Uhat_\perp\tps U}{2}^2} \quad\mbox{(by definition of $L_2$ matrix norm)}\\
\end{align*}
Therefore,
\begin{align}
\label{eq:lem_HKZ_9_misc1} \pnorm{\Uhat\tps U x}{2} &\geq \pnorm{x}{2}\sqrt{1 - \pnorm{\Uhat_\perp\tps U}{2}^2}
\end{align}
Note that
\begin{align*}
\pnorm{\Uhat_\perp\tps U}{2}^2 &\leq  \etwoone^2/\sPtwoone^2 \quad\mbox{(by Corollary~\ref{cor_HKZ_22})}\\
&\leq \frac{\etwoone^2}{(1-\varepsilon)^2\sPtwoone^2} \quad\mbox{(since $0 \leq \varepsilon < 1/2$)} \\
&= \varepsilon_0\quad\mbox{(by definition)}
\end{align*}
Hence, by combining the above with equation~\eqref{eq:lem_HKZ_9_misc1}, since $0 \leq \varepsilon_0 < 1$:
\begin{align}
\label{eq:lem_HKZ_9_misc2} \pnorm{\Uhat\tps Ux}{2} &\geq \pnorm{x}{2}\sqrt{1 - \varepsilon_0}\quad\mbox{(for all $x\in\Rbbvec{k}$)}
\end{align}
The remaining claims follow by taking different choices of $x$ in equation~\eqref{eq:lem_HKZ_9_misc2}, and by using the intuition that the smallest singular value of a matrix is the smallest possible $L_2$ norm of a unit-length vector after the matrix has left-multiplied that vector, and the particular vector for which this holds is the corresponding right singular vector.
For claim 3, let $x = \Sigma V\tps \nuvec_k\{\Uhat\tps\Ptwoone\}$. Then by equation~\eqref{eq:lem_HKZ_9_misc2}:

\begin{align*}
\pnorm{\Uhat\tps U\Sigma V\tps \nuvec_k\{\Uhat\tps\Ptwoone\}}{2} &\geq \pnorm{\Sigma V\tps \nuvec_k\{\Uhat\tps\Ptwoone\}}{2}\sqrt{1 - \varepsilon_0}
\end{align*}

Since $\Ptwoone = U\Sigma V\tps$, and $\pnorm{\Sigma V\tps \nuvec_k\{\Sigma V\tps\}}{2} \leq \pnorm{\Sigma V\tps \nuvec_k\{\Uhat\tps\Ptwoone\}}{2}$ by definition of $\nuvec_k\{\Sigma V\tps\}$, we have:

\begin{align*}
\pnorm{\Uhat\tps \Ptwoone \nuvec_k\{\Uhat\tps\Ptwoone\}}{2} &\geq \pnorm{\Sigma V\tps \nuvec_k\{\Sigma V\tps\}}{2}\sqrt{1 - \varepsilon_0}  \\
\sigma_k(\Uhat\tps\Ptwoone) &\geq \sigma_k(\Sigma V\tps) \sqrt{1 - \varepsilon_0}\quad\mbox{(by definition of $\sigma_k(\Uhat\tps\Ptwoone),\sigma_k(\Sigma V\tps)$)} \\
\sigma_k(\Uhat\tps\Ptwoone) &\geq \sigma_k(\Ptwoone) \sqrt{1 - \varepsilon_0}\quad\mbox{($\because \sigma_k(\Sigma V\tps) = \sigma_k(\Ptwoone)$)}
\end{align*}

which proves claim 3.

For claim 4, first recall that $OR$ can be exactly expressed as $\Ptwoone(S\diag(\vec{\pi})O\tps )^+$ (equation~\eqref{eq:OR_P21}). For brevity, let $\mathcal{A} = (S\diag(\vec{\pi})O\tps )^+$, so that $OR = \Ptwoone \mathcal{A}$. Then, let $x = \Sigma V\tps \mathcal{A} \nuvec_k\{\Uhat\tps OR\}$ in equation~\eqref{eq:lem_HKZ_9_misc2}:

\begin{align*}
\pnorm{\Uhat\tps U\Sigma V\tps \mathcal{A} \nuvec_k\{\Uhat\tps OR\}}{2} &\geq \pnorm{\Sigma V\tps \mathcal{A} \nuvec_k\{\Uhat\tps OR\}}{2}\sqrt{1 - \varepsilon_0}
\end{align*}

Since $\Ptwoone = U\Sigma V\tps$, and $\pnorm{\Sigma V\tps \mathcal{A} \nuvec_k\{\Sigma V\tps \mathcal{A}\}}{2} \leq  \pnorm{\Sigma V\tps \mathcal{A} \nuvec_k\{\Uhat\tps OR\}}{2}$ by definition of $\nuvec_k\{\Sigma V\tps \mathcal{A}\}$, we get:

\begin{align*}
\pnorm{\Uhat\tps \Ptwoone \mathcal{A} \nuvec_k\{\Uhat\tps OR\}}{2} &\geq \pnorm{\Sigma V\tps \mathcal{A} \nuvec_k\{\Sigma V\tps \mathcal{A}\}}{2}\sqrt{1 - \varepsilon_0} \\
\pnorm{\Uhat\tps OR \nuvec_k\{\Uhat\tps OR\}}{2} &\geq \sigma_k(\Sigma V\tps \mathcal{A}) \sqrt{1 - \varepsilon_0}\quad\mbox{(by equation~\eqref{eq:OR_P21})}
\end{align*}

By definition of $\sigma_k(\Uhat\tps OR),\sigma_k(\Sigma V\tps \mathcal{A})$, we see that

\begin{align*}
\sigma_k(\Uhat\tps OR) &\geq \sigma_k(\Sigma V\tps \mathcal{A}) \sqrt{1 - \varepsilon_0} \\
\sigma_k(\Uhat\tps OR) &\geq \sigma_k(OR) \sqrt{1 - \varepsilon_0}\quad\mbox{($\because \sigma_k(\Sigma V\tps \mathcal{A}) = \sigma_k(\Ptwoone \mathcal{A}) = \sigma_k(OR)$)}
\end{align*}

hence proving claim 4.
\end{proof} \\

Define the following observable representation using $U=\Uhat$ , which constitutes a \textit{true} observable representation for the HMM as long as $\UOR$ is invertible:
\begin{align*}
\btilde_\infty &= (\Ptwoone\tps \Uhat)^+\Pone = \UhOR ^{-T}R\tps \vec{1}_m \\
\Btilde_x &= (\Uhat\tps \Pthreexone)(\Uhat\tps \Ptwoone)^+ = \UhOR W_x\UhORi \quad \mbox{for $x = 1,\ldots,n$} \\
\btilde_1 &= \Uhat\tps \Pone
\end{align*}
Define the following error measures of estimated parameters with respect to the true observable representation. The error vector in $\delta_1$ is projected to $\Rbbvec{m}$ before applying the vector norm, for convenience in later theorems.
%
\begin{align*}
\delta_\infty &= \pnorm{(\Uhat\tps O)\tps (\bhat_\infty - \btilde_\infty)}{\infty}  \\
\Delta_x &= \pnorm{\UhORi\left(\Bhatx - \Btilde_x\right)\UhOR}{1} = \pnorm{\UhORi\Bhatx\UhOR  - W_x}{1} \\
\Delta &= \sum_{x}\Delta_x \\
\delta_1 &= \pnorm{R\UhORi(\bhat_1 - \btilde_1)}{1} = \pnorm{R\UhORi\bhat_1 - \vec{\pi}}{1}
\end{align*}
The next Lemma proves that the estimated parameters $\bhinf,\Bhatx,\bhone$ are close to the true parameters $\btinf, \Btildex,\btone$ if the sampling errors $\eone, \etwoone, \ethreexone$ are small:
\begin{lemma}\label{lem_HKZ_10}[Modification of HKZ Lemma 10] Assume $\etwoone < \sPtwoone/3$. Then: \\
\begin{align*}
\delta_\infty &\leq 4\cdot\left(\frac{\etwoone}{\sPtwoone^2} + \frac{\eone}{3\sPtwoone}\right) \\
\Delta_x &\leq \frac{8}{\sqrt{3}}\cdot\frac{\sqrt{k}}{\sOR}\cdot\left(\Pr[x_2 = x]\cdot\frac{\etwoone}{\sPtwoone^2} + \frac{\Sigma_x \ethreexone}{3\sPtwoone}\right) \\
\Delta &\leq \frac{8}{\sqrt{3}}\cdot\frac{\sqrt{k}}{\sOR}\cdot\left(\frac{\etwoone}{\sPtwoone^2} + \frac{\Sigma_x \ethreexone}{3\sPtwoone}\right) \\
\delta_1 &\leq \frac{2}{\sqrt{3}}\cdot\frac{\sqrt{k}}{\sOR}\cdot\eone
\end{align*}
\end{lemma}
\begin{proof} Note that the assumption on $\etwoone$ guarantees $\UhOR$ to be invertible by Lemma~\ref{lem_HKZ_9}, claim 4.

\paragraph{$\delinf$ bound:} We first see that $\delinf$ can be bounded by $\pnorm{\bhinf - \btinf}{2}$:
\begin{align*}
\delinf &= \infnorm{(O\tps U)(\bhinf - \btinf)} \\
&\leq \infnorm{O\tps}\infnorm{U(\bhinf - \btinf)} \\
&\leq \infnorm{U(\bhinf - \btinf)} \\
&\leq \pnorm{U(\bhinf - \btinf)}{2} \\
&\leq \pnorm{\bhinf - \btinf}{2} \\
\end{align*}
In turn, this leads to the following expression:
\begin{align*}
\pnorm{\bhinf - \btinf}{2} &= \pnorm{(\Phtwoone\tps\Uhat)^+\Phone - (\Ptwoone\tps\Uhat)^+\Pone}{2} \\
&= \pnorm{(\Phtwoone\tps\Uhat)^+\Phone - (\Ptwoone\tps\Uhat)^+\Phone + (\Ptwoone\tps\Uhat)^+\Phone - (\Ptwoone\tps\Uhat)^+\Pone}{2} \\
&= \pnorm{\left((\Phtwoone\tps\Uhat)^+ - (\Ptwoone\tps\Uhat)^+\right)\Phone + (\Ptwoone\tps\Uhat)^+(\Phone - \Pone)}{2} \\
&\leq \pnorm{((\Phtwoone\tps\Uhat)^+ - (\Ptwoone\tps\Uhat)^+)\Phone}{2} + \pnorm{(\Ptwoone\tps\Uhat)^+(\Phone - \Pone)}{2} \\
&\leq \pnorm{((\Phtwoone\tps\Uhat)^+ - (\Ptwoone\tps\Uhat)^+)}{2}\pnorm{\Phone}{1} + \pnorm{(\Ptwoone\tps\Uhat)^+}{2}\pnorm{(\Phone - \Pone)}{2}
\end{align*}
The last step above obtains from the consistency of the $L_2$ matrix norm with $L_1$ vector norm, and from the definition of $L_2$ matrix norm (spectral norm) as $\pnorm{A}{2} = \max \frac{\pnorm{Ax}{2}}{\pnorm{x}{2}}$. Now, recall that $\Uhat$ has orthonormal columns, and hence multiplying a matrix with $\Uhat$ cannot increase its spectral norm. Hence,
\[ \pnorm{\Phtwoone\tps\Uhat - \Ptwoone\tps\Uhat }{2} = \pnorm{(\Phtwoone\tps - \Ptwoone\tps)\Uhat }{2} \leq \pnorm{\Phtwoone\tps - \Ptwoone\tps}{2} = \etwoone. \]
So, we can use Lemma~\ref{lem_HKZ_23} to bound the $L_2$-distance between pseudoinverses of $\Phtwoone\tps\Uhat$ and $\Ptwoone\tps\Uhat$ using $\etwoone$ as an upper bound on the difference between the matrices themselves. Also recall that singular values of the pseudoinverse of a matrix are the reciprocals of the matrix singular values. Substituting this in the above expression, along with the facts that $\sigma_k(\Phtwoone\tps\Uhat) = \sigma_k(\Phtwoone)$, $\pnorm{\Phone}{1} = 1$ and $\pnorm{(\Phone - \Pone)}{2} = \eone$, gives us:
\begin{align*}
\pnorm{\bhinf - \btinf}{2} &\leq \frac{1+\sqrt{5}}{2}\cdot\frac{\etwoone}{\min\left(\shPtwoone,\sigma_k(\Ptwoone\tps\Uhat)\right)^2} + \frac{\eone}{\sigma_k(\Ptwoone\tps \Uhat)}
\end{align*}
Now, to simplify the last expression further, consider Lemma~\ref{lem_HKZ_9} in the above context. Here, $\etwoone \leq \sPtwoone/3$ and hence $\varepsilon = 1/3$. Therefore $\shPtwoone = \sigma_k(\Uhat\tps\Phtwoone) \geq (2/3)\sPtwoone$ and $\sigma_k(\Uhat\tps\Ptwoone) \geq \sqrt{1-\varepsilon_0}\sPtwoone$. Hence
\begin{align*}
\min\left(\shPtwoone,\sigma_k(\Ptwoone\tps\Uhat)\right)^2 &= \sPtwoone^2\cdot\min(2/3,\sqrt{1-\varepsilon_0})^2
\end{align*}
The latter term is larger since
\begin{align*}
\varepsilon_0 &= \frac{\etwoone^2}{((1-\varepsilon)\sPtwoone)^2} \\
&\leq \frac{\sPtwoone^2/9}{4\sPtwoone^2/9} \\
&= 1/4 \\
\Rightarrow \sqrt{1-\varepsilon_0} &\geq \sqrt{3}/2 > 2/3
\end{align*}
Therefore $\min\left(\shPtwoone,\sigma_k(\Ptwoone\tps\Uhat)\right)^2 \geq \sPtwoone^2(2/3)^2$. Plugging this into the expression above along with the fact that $\sigma_k(\Uhat\tps \Ptwoone) \geq (\sqrt{3}/2)\sPtwoone$, we prove the required result for $\delinf$:
\begin{align*}
\delinf &\leq \frac{1+\sqrt{5}}{2}\cdot\frac{9\etwoone}{4\sPtwoone^2} + \frac{2\eone}{\sqrt{3}\sPtwoone} \\
&\leq 4\cdot\left(\frac{\etwoone}{\sPtwoone^2} + \frac{\eone}{\sPtwoone}\right)
\end{align*}

\paragraph{$\Delta_x$,$\Delta$ bounds:}. We first bound each term $\Delta_x$ by $\sqrt{k}\pnorm{\Bhatx - \Btildex}{2}/\sigma_k(\UhOR)$:
\begin{align*}
\Delta_x &= \pnorm{\UhORi\left(\Bhatx - \Btilde_x\right)\UhOR}{1}  \\
&\leq \pnorm{\UhORi(\Bhatx - \Btildex)\Uhat\tps}{1}\pnorm{OR}{1} \quad\mbox{(by norm consistency)}\\
&\leq \sqrt{k}\pnorm{\UhORi(\Bhatx - \Btildex)\Uhat\tps}{2}\pnorm{OR}{1} \quad\mbox{(by $L_1$ vs.\@ $L_2$ norm inequality)}\\
&\leq \sqrt{k}\pnorm{\UhORi}{2}\pnorm{\Bhatx - \Btildex}{2}\pnorm{\Uhat\tps}{2}\pnorm{O}{1}\pnorm{R}{1} \quad\mbox{(by norm consistency)}\\
&\leq \sqrt{k}\pnorm{\UhORi}{2}\pnorm{\Bhatx - \Btildex}{2} \quad\mbox{$\left(\pnorm{\Uhat\tps}{2},\pnorm{O}{1},\pnorm{R}{1} \leq 1\right)$}\\
&= \sqrt{k}\pnorm{\Bhatx - \Btildex}{2}/\sigma_k\UhOR \quad\mbox{($\because \sigma_{\max}\UhORi = 1/\sigma_{\min}\UhOR$)}
\end{align*}
The term $\pnorm{\Bhatx - \Btildex}{2}$ in the numerator can be bounded by
\begin{align*}
\pnorm{\Bhatx - \Btildex}{2} &= \pnorm{(\Uhat\tps\Pthreexone)(\Uhat\tps\Ptwoone)^+ - (\Uhat\tps\Phthreexone)(\Uhat\tps\Phtwoone)^+}{2} \\
&\leq \pnorm{(\Uhat\tps\Pthreexone)\left((\Uhat\tps\Ptwoone)^+ - (\Uhat\tps\Phtwoone)^+\right)}{2} + \pnorm{\Uhat\tps\left(\Pthreexone - \Phthreexone\right)(\Uhat\tps\Ptwoone)^+}{2} \\
&\leq \pnorm{\Pthreexone}{2}\cdot\frac{1+\sqrt{5}}{2}\cdot\frac{\etwoone}{\min\left(\shPtwoone,\sigma_k(\Uhat\tps\Ptwoone)\right)^2} + \frac{\ethreexone}{\sigma_k(\Uhat\tps\Ptwoone)} \\
&\leq \Pr[x_2 = x]\cdot\frac{1+\sqrt{5}}{2}\cdot\frac{\etwoone}{\min\left(\shPtwoone,\sigma_k(\Uhat\tps\Ptwoone)\right)^2} + \frac{\ethreexone}{\sigma_k(\Uhat\tps\Ptwoone)}
\end{align*}
where the second inequality is from Lemma~\ref{lem_HKZ_23} and the last one uses the fact that
\[\pnorm{\Pthreexone}{2} \leq \pnorm{\Pthreexone}{F} = \sqrt{\sum_{i,j}[\Pthreexone]^2_{i,j}} \leq \sum_{i,j} [\Pthreexone]_{i,j} = \Pr[x_2 = x].\]
Applying Lemma~\ref{lem_HKZ_9} as in the $\delinf$ bound above, gives us the required result on $\Delta_x$. Summing both sides over $x$ results in the required bound on $\Delta$.

\paragraph{$\delone$ bound:}
For $\delta_1$, we invoke Condition~\ref{con_R_L1} to use the fact that $\pnorm{R}{1} \leq 1$. Specifically,
\begin{align*}
\delta_1 &= \pnorm{R\UhORi\Uhat\tps(\Phone - \Pone)}{1}\\
&\leq \pnorm{R}{1}\pnorm{\UhORi\Uhat\tps(\Phone - \Pone)}{1} \quad \text{(norm consistency)} \\
&\leq \sqrt{k}\pnorm{R}{1}\pnorm{\UhORi\Uhat\tps(\Phone - \Pone)}{2} \quad \text{($\pnorm{x}{1} \leq \sqrt{n} \pnorm{x}{2}$ for any $x\in\Rbbvec{n})$} \\
&\leq \sqrt{k}\pnorm{R}{1}\pnorm{\UhORi\Uhat\tps}{2}\pnorm{(\Phone - \Pone)}{2} \quad \text{(norm consistency)}\\
&\leq  \sqrt{k}\pnorm{R}{1}\pnorm{\UhORi}{2}\cdot\eone  \quad \text{(defn.\@ of $\eone$, $\Uhat\tps$ has orthogonal columns) }\\
&=  \frac{\sqrt{k}\eone}{\sigma_k \UhOR} \quad \text{($\pnorm{R}{1} \leq 1$, defn.\@ of $L_2$-norm) }
\end{align*}
The desired bound on $\delta_1$ is obtained by using Lemma~\ref{lem_HKZ_9}. With $\varepsilon,\varepsilon_0$ as described in the above proof for $\delinf$, we have that $\sigma_k \UhOR \geq (\sqrt{3}/2)\sigma_k\UOR$. The required bound follows by plugging this inequality into the above upper bound for $\delta_1$.
\end{proof} \\

\subsection{Proof of Theorem~\ref{thm_HKZ_6}}
\label{RRHMM-sec-proofs-proof1}
The following Lemmas~\ref{lem_HKZ_11} and \ref{lem_HKZ_12} together with Lemmas \ref{lem_HKZ_8},\ref{lem_HKZ_9},\ref{lem_HKZ_10} above, constitute the proof of Theorem~\ref{thm_HKZ_6} on joint probability accuracy. We state the results based on appropriate modifications of HKZ, and provide complete proofs. We also describe how the proofs generalize to the case of handling continuous observations using Kernel Density Estimation (KDE). First, define the following as in HKZ
\begin{align*}
\epsilon(i) &= \min\left\{\sum_{j\in S} \Pr[x_2 = j] : S \subseteq \{1 \ldots n\},\abs{S} = n - i \right\}
\end{align*}
and let
\begin{align*}
n_0(\varepsilon) &= \min\{ i : \epsilon(i) \leq \varepsilon\}
\end{align*}
The term $n_0(\varepsilon)$, which occurs in the theorem statement, can be interpreted as the minimum number of discrete observations that accounts for $1-\epsilon$ of total marginal observation probability mass. Since this can be much lower than (and independent of) $n$ in many applications, the analysis of HKZ is able to use $n_0$ instead of $n$ in the sample complexity bound. This is useful in domains with large $n$, and our relaxation of HKZ preserves this advantageous property.

The following lemma quantifies how estimation errors accumulate while computing the joint probability of a length $t$ sequence, due to errors in $\Bhatx$ and $\bhat$.
\begin{lemma}\label{lem_HKZ_11}[Modification of HKZ Lemma 11] Assume $\Uhat\tps OR$ is invertible. For any time $t$:
\[ \sum_{x_{1:t}}\pnorm{R\UORi\left(\Bhat_{x_{t:1}}\bhat_1 - \Btilde_{x_{t:1}}\btilde_1\right)}{1} \leq (1+\Delta)^t \delta_1 + (1 + \Delta)^t  - 1\]
\end{lemma}
\begin{proof}
Proof by induction. The base case for $t=0$, i.e.\@ that $\pnorm{R\UORi(\bhat_1 - \btilde_1)}{1} \leq \delta_1$ is true by definition of $\delta_1$. For the rest, define unnormalized states $\bhat_t = \bhat_t(x_{1:t-1}) = \Bhat_{x_t-1:1}\bhat_1$ and $\btilde_t = \btilde_t(x_{1:t-1}) = \Btilde_{x_{t-1}:1}\btilde_1$. For some particular $t >1$, assume the inductive hypothesis as follows

\[ \sum_{x_{1:t}}\pnorm{R\UhORi\left(\bhat_t - \btilde_t\right)}{1} \leq (1+\Delta)^t \delta_1 + (1 + \Delta)^t  - 1\]

The sum over $x_{1:t}$ in the LHS can be decomposed as:
\begin{align*}
&\sum_{x_{1:t}}\pnorm{R\UhORi\left(\bhat_t - \btilde_t\right)}{1} \\
&= \quad \sum_{x}\sum_{x_{1:t-1}}\pnorm{R\UORi\left((\Bhat_{x_t} - \Btilde_{x_t})\btilde_t + (\Bhat_{x_t} - \Btilde_{x_t})(\bhat_t - \btilde_t) + \Btilde_{x_t}(\bhat_t - \btilde_t)\right)}{1}
\end{align*}
Using triangle inequality, the above sum is bounded by
\begin{align*}
&\sum_{x_t} \sum_{x_{1:t-1}} \pnorm{R\UhORi\left(\Bhat_{x_t} - \Btilde_{x_t}\right)(\Uhat\tps O)}{1}\pnorm{R\UhORi\btilde_t}{1} \\
&+ \sum_{x_t} \sum_{x_{1:t-1}} \pnorm{R\UhORi\left(\Bhat_{x_t} - \Btilde_{x_t}\right)(\Uhat\tps O)}{1}\pnorm{R\UhORi\left(\bhat_t - \btilde_t\right)}{1} \\
&+ \sum_{x_t} \sum_{x_{1:t-1}} \pnorm{R\UhORi \Btilde_t\UhOR\UhORi\left(\bhat_t - \btilde_t\right)}{1} \\
\end{align*}
Each of the above double sums is bounded separately. For the first, we note that $\pnorm{R\UhORi \btilde_t}{1} = \Pr[x_{1:t-1}]$, which sums to $1$ over $x_{1:t-1}$. The remainder of the double sum is bounded by $\Delta$, by definition. For the second double sum, the inner sum over $\pnorm{R\UhORi (\bhat_t - \btilde_t)}{1}$ is bounded using the inductive hypothesis. The outer sum scales this bound by $\Delta$, by definition. Hence the second double sum is bounded by $\Delta((1 + \Delta)^{t-1}\delta_1 + (1 + \Delta)^{t-1} - 1)$. Finally, we deal with the third double sum as follows. We first replace $\UhORi\Btilde_t\UhOR$ by $W_{x_t}$, and note that $R\cdot W_{x_t} = A_{x_t} R$. Since $A_{x_t}$ is entry-wise nonnegative by definition, $\pnorm{A_{x_t}\vvec}{1} \leq \onevect_m A_{x_t} |\vvec|$, where $|\vvec|$ denotes element-wise absolute value. Also note that $\onevect_m \sum_{x_t} A_{x_t}|\vvec| = \onevect_m T |\vvec| = \onevect_m |\vvec| = \pnorm{\vvec}{1}$. Using this result with $\vvec = R\UhORi(\bhat_t - \btilde_t)$ in the third double sum above, the inductive hypothesis bounds the double sum by $(1 + \Delta)^{t-1}\delone + (1 + \Delta)^{t-1} - 1$. Combining these three bounds gives us the required result:
\begin{align*}
&\sum_{x_{1:t}}\pnorm{R\UhORi\left(\bhat_t - \btilde_t\right)}{1} \\
&\leq \Delta + \Delta((1 + \Delta)^{t-1}\delta_1 + (1 + \Delta)^{t-1} - 1)  + (1 + \Delta)^{t-1}\delone + (1 + \Delta)^{t-1} - 1 \\
&= \Delta + (1+\Delta)((1 + \Delta)^{t-1}\delta_1 + (1 + \Delta)^{t-1} - 1) \\
&= \Delta + (1 + \Delta)^{t}\delta_1 + (1 + \Delta)^{t} - 1 - \Delta \\
&= (1 + \Delta)^{t}\delta_1 + (1 + \Delta)^{t} - 1
\end{align*}
thus completing the induction.
\end{proof}

The following lemma bounds the effect of errors in the normalizer $\bhinf$.
\begin{lemma}\label{lem_HKZ_12}[Modification of HKZ Lemma 12] Assume $\etwoone \leq \sPtwoone/3$. Then for any $t$,
\[ \sum_{x_{1:t}}\left|\Pr[x_{1:t}] - \widehat{\Pr}[x_{1:t}]\right| \leq  (1+\delinf)(1+\delone)(1+\Delta)^t - 1\]
\end{lemma}
\begin{proof}
First note that the upper bound on $\etwoone$ along with Lemma~\ref{lem_HKZ_9}, ensure that $\sigma_k \UhOR > 0$ and so $\UhOR$ is invertible. The LHS above can be decomposed into three sums that are dealt with separately:
\begin{align*}
\sum_{x_{1:t}}\left|\Pr[x_{1:t}] - \widehat{\Pr}[x_{1:t}]\right| &= \sum_{x_{1:t}}\left|\bhinft \Bhat_{x_{t:1}}\bhone - \binft B_{x_{t:1}}\bone\right| \\
&= \sum_{x_{1:t}} \abs{\bhinft \Bhat_{x_{t:1}}\bhone - \btinft \Btilde_{x_{t:1}}\btone} \\
&\leq \sum_{x_{1:t}} \abs{(\bhinf - \btinf)\tps\UhOR\UhORi\Btilde_{x_{t:1}}\btone} \\
&\quad + \sum_{x_{1:t}} \abs{(\bhinf - \btinf)\tps\UhOR\UhORi(\Bhat_{x_{t:1}}\bhone - \Btilde_{x_{t:1}}\btone)} \\
&\quad + \sum_{x_{1:t}} \abs{\btinft\UhOR\UhORi(\Bhat_{x_{t:1}}\bhone - \Btilde_{x_{t:1}}\btone)}
\end{align*}

The first sum can be bounded as follows, using H\"{o}lders inequality and bounds from Lemma~\ref{lem_HKZ_10}:
\begin{align*}
\sum_{x_{1:t}} \abs{(\bhinf - \btinf)\tps\UhOR\UhORi\Btilde_{x_{t:1}}\btone} &\leq \sum_{x_{1:t}} \infnorm{(\Uhat\tps O)\tps (\bhinf - \btinf)} \pnorm{R\UhORi\Btilde_{x_{t:1}}\btone}{1} \\
& \leq \sum_{x_{1:t}} \delinf \pnorm{A_{x_{t:1}}\pivec}{1} \\
&= \sum_{x_{1:t}} \delinf \Pr[x_{1:t}] \\
&= \delinf
\end{align*}
The second sum can be bounded also using H\"{o}lders, as well as the bound in Lemma~\ref{lem_HKZ_11}:
\begin{align*}
&\sum_{x_{1:t}} \abs{(\bhinf - \btinf)\tps\UhOR\UhORi(\Bhat_{x_{t:1}}\bhone - \Btilde_{x_{t:1}}\btone)}  \\
&\qquad\leq \infnorm{(\Uhat\tps O)\tps(\bhinf - \btinf)}\pnorm{R\UhORi (\Bhat_{x_{t:1}}\bhone - \Btilde_{x_{t:1}}\btone)}{1} \\
&\qquad\leq \delinf((1+\Delta)^t\delone + (1 + \Delta)^t -1)
\end{align*}
The third sum again uses Lemma~\ref{lem_HKZ_11}:
\begin{align*}
\sum_{x_{1:t}} \abs{\btinft\UhOR\UhORi(\Bhat_{x_{t:1}}\bhone - \Btilde_{x_{t:1}}\btone)} &= \sum_{x_{1:t}} \abs{\onevect R\UhORi(\Bhat_{x_{t:1}}\bhone - \Btilde_{x_{t:1}}\btone)} \\
&\leq \pnorm{R\UhORi(\Bhat_{x_{t:1}}\bhone - \Btilde_{x_{t:1}}\btone)}{1} \\
&\leq (1+\Delta)^t\delone + (1 + \Delta)^t -1
\end{align*}

Adding these three sums gives us:
\begin{align*}
\sum_{x_{1:t}}\left|\Pr[x_{1:t}] - \widehat{\Pr}[x_{1:t}]\right| &\leq \delinf + \delinf((1+\Delta)^t\delone + (1 + \Delta)^t -1) +  (1+\Delta)^t\delone + (1 + \Delta)^t -1\\
&\leq \delinf + (1 + \delinf)((1+\Delta)^t\delone + (1 + \Delta)^t -1)
\end{align*}
which is the required bound.
\end{proof}

\begin{proof}(Theorem~\ref{thm_HKZ_6}).  Assume $N$ and $\varepsilon$ as in the theorem statement:

\begin{align*}
\varepsilon &= \sigma_k(OR) \sigma_k(\Ptwoone)\epsilon/(4t\sqrt{k}) \\
N &\geq C \cdot \frac{t^2}{\epsilon^2} \cdot \left(\frac{k}{\sigma_k(OR)^2 \sigma_k(P_{2,1})^4} + \frac{k\cdot n_0(\varepsilon)}{\sigma_k(OR)^2 \sigma_k (P_{2,1})^2}  \right)\cdot \log (1/\eta)
\end{align*}

 First note that
\[\sum_{x_{1:t}}\left|\Pr[x_{1:t}] - \widehat{\Pr}[x_{1:t}]\right| \leq 2\]
since it is the $L_1$ difference between two stochastic vectors. Therefore, the theorem is vacuous for $\epsilon \geq 2$. Hence we can assume
\[\epsilon < 1\]
in the proof and let the constant $C$ absorb the factor $4$ difference due to the $1/\epsilon^2$ term in the expression for $N$.

The proof has three steps. We first list these steps then prove them below.

\underline{First step}: for a suitable constant $C$, the following sampling error bounds follow from Lemma~\ref{lem_HKZ_8}:
\begin{subequations} \label{eq:epsmins}
\begin{align}
\eone &\leq \min \left(.05\cdot(3/8)\cdot\sPtwoone \cdot\epsilon, .05\cdot(\sqrt{3}/2)\cdot\sOR \cdot(1/\sqrt{k})\cdot \epsilon\right) \\
\etwoone &\leq \min \left(.05\cdot(1/8)\cdot \sPtwoone ^2\cdot(\epsilon/5), .01\cdot(\sqrt{3}/8)\cdot\sOR \cdot \sigma_k (\Ptwoone)^2\cdot(1/(t\sqrt{k}))\cdot \epsilon \right) \\
\sum_x \epsilon_{3,x,1} &\leq 0.39 \cdot (3\sqrt{3}/8)\cdot\sOR \cdot\sPtwoone \cdot(1/(t\sqrt{k}))\cdot\epsilon
\end{align}
\end{subequations}
\underline{Second step}: Lemma~\ref{lem_HKZ_10} together with equations~\eqref{eq:epsmins} imply:
\begin{subequations}\label{eq:delbounds}
\begin{align}
\delinf &\leq .05\epsilon \\
\delone &\leq .05\epsilon \\
\Delta &\leq 0.4\epsilon/t
\end{align}
\end{subequations}
\underline{Third step}: By Lemma~\ref{lem_HKZ_12}, equations~\eqref{eq:delbounds} and the inequality
\begin{align}
\label{eq:a_by_t_ineq} (1+(a/t))^t &\leq 1 + 2a \quad \mbox{for $a \leq 1/2$}
\end{align}
we get the theorem statement.

\underline{Proof of first step}: Note that for any value of matrix $\Ptwoone$, we can upper-bound $\sPtwoone$ by $1$:
\begin{align*}
\sPtwoone &\leq \sigma_1(\Ptwoone) \\
&= \max_{\pnorm{x}{2} = 1} \pnorm{\Ptwoone x}{2} \\
&= \max_{\pnorm{x}{2} = 1} \left( \sum_{j=1}^{n} \left( \sum_{i=1}^{n}[\Ptwoone]_{ij} x_i \right)^2 \right)^{1/2} \\
&\leq  \max_{\pnorm{x}{2} = 1} \sum_{j=1}^{n} \left| \sum_{i=1}^{n}[\Ptwoone]_{ij} x_i \right| \quad\mbox{(by norm inequality)} \\
&\leq \sum_{j=1}^{n}\sum_{i=1}^{n} \left| [\Ptwoone]_{ij}\right| \quad\mbox{($|x_i| \leq 1$ since $\pnorm{x}{2} = 1$)} \\
&=  \sum_{j=1}^{n}\sum_{i=1}^{n} [\Ptwoone]_{ij} \quad\mbox{(by non-negativity of $\Ptwoone$)} \\
&= 1 \quad\mbox{(by definition)}
\end{align*}
Similarly, for any column-stochastic observation probability matrix $O$ we can bound $\sOR$ by $\sqrt{k}$. First see that $\sigma_1(O)\leq \sqrt{m}$:
\begin{align*}
\sigma_1(O) &= \max_{\pnorm{x}{2} = 1} \pnorm{Ox}{2} \\
&= \max_{\pnorm{x}{2} = 1} \left( \sum_{j=1}^m \sum_{i=1}^n (O_{ij} x_i)^2 \right)^{1/2}  \\
&\leq \max_{\pnorm{x}{2} = 1} \left( \sum_{j=1}^m \sum_{i=1}^n O_{ij}^2 \right)^{1/2} \quad\mbox{($\pnorm{x}{2} = 1 \Rightarrow |x_i| \leq 1$)} \\
&\leq \left( \sum_{j=1}^m  (\sum_{i=1}^n O_{ij})^2 \right)^{1/2} \quad\mbox{(by triangle inequality)}\\
&\leq \left( \sum_{j=1}^m 1^2 \right)^{1/2} \quad\mbox{(by definition of $O$)} \\
&= \sqrt{m}
\end{align*}
Now the bound on $\sOR$ follows from Condition~\ref{con_R_unif} i.e.\@ $\sigma_k(OR) \leq \sqrt{k/m}$:
\begin{align*}
\sOR &= \min_{\pnorm{x}{2} = 1} \pnorm{ORx}{2} \\
&\leq \pnorm{O}{2}\cdot\min_{\pnorm{x}{2} = 1} \pnorm{Rx}{2} \quad\mbox{(by norm consistency)}\\
&\leq \sqrt{m}\min_{\pnorm{x}{2} = 1}\sqrt{\sum_{i=1}^m \sum_{j=1}^k (R_{ij}x_j)^2} \quad\mbox{($\because \pnorm{A}{2} = \sigma_1(A)$ for any matrix $A$)}\\
\end{align*}
Assume the $c^{th}$ column of $R$ obeys Condition~\ref{con_R_unif} for some $1\leq c \leq k$. Also assume $x = e_c$, the $c^{th}$ column of the $k\times k$ identity matrix, which obeys the constraint $\pnorm{x}{2} = 1$. Then every component of the inner sum is zero except when $j = c$, and the $\min$ expression can only get larger:
\begin{align*}
\sOR &\leq \sqrt{m}\sqrt{\sum_{i=1}^m R_{ic}^2} \\
&= \sqrt{m}\pnorm{R[\cdot,c]}{2} \\
&\leq \sqrt{m}\sqrt{k/m} \\
&= \sqrt{k}
\end{align*}
hence proving that $\sOR \leq \sqrt{k}$.

Now we begin the proof with the $\eone$ case. Choose a $C$ that satisfies all previous bounds and also obeys $(\sqrt{C}/4)\cdot 0.05\cdot(3/8)  \geq 1$.
\begin{align}
\label{eq:samplebound_step1}
\eone &\leq \sqrt{1/N}(\sqrt{\ln (3/\eta)} + 1) \quad\mbox{(by Lemma~\ref{lem_HKZ_8})} \\
&\leq \sqrt{1/N}(2\sqrt{\ln (3/\eta)}) \quad\mbox{(since $\sqrt{\ln (3/\eta)} \geq \sqrt{\ln 3} > 1$)}
\end{align}
Now, plugging in the assumed value of $N$:
\begin{align}
\label{eq:eone_ineq_with_N} \eone &\leq \frac{2\epsilon (\sPtwoone^2\sOR)}{t\sqrt{Ck(1+n_0(\varepsilon)\sPtwoone^2)}}\sqrt{\frac{\ln (3/\eta)}{\ln (1/\eta)}}
\end{align}
Any substitutions that increase the right hand side of the above inequality preserve the inequality. We now drop the additive $1$ in the denominator, replace $\sqrt{\ln (3/\eta)/\ln(1/\eta)}$ by $2$ since it is at most $\sqrt{\ln 3}$,  and drop the factors $t,\sqrt{n_0(\varepsilon)}$ from the denominator.
\begin{align}
\label{eq:samplebound_step2}
\eone &\leq \frac{4 \sOR\sPtwoone^2 \epsilon}{\sqrt{Ck}\sPtwoone} \\
&\leq \frac{1}{\sqrt{C}}\cdot 4\cdot \left[\sOR/\sqrt{k}\right]\cdot\left[\sPtwoone\right]\cdot\epsilon \nonumber\\
&\leq \frac{1}{\sqrt{C}}\min\left(4\cdot\sPtwoone\cdot\epsilon, 4\cdot\sOR\cdot 1/\sqrt{k}\cdot\epsilon\right) \quad\mbox{($\because$ both $\left[\sOR/\sqrt{k}\right]$ and $\left[\sPtwoone\right]$ are $\leq 1$)} \nonumber\\
&= \frac{1}{C'}\min\left(0.05\cdot 3/8\cdot\sPtwoone\cdot\epsilon, 0.05\cdot \sqrt{3}/2 \cdot \sOR\cdot 1/\sqrt{k} \cdot\epsilon\right) \nonumber\\ &\quad\mbox{(for $C' = \frac{\sqrt{C}}{4}\cdot 0.05\cdot 3/8$)} \nonumber\\
&= \min\left(0.05\cdot 3/8 \cdot \sPtwoone \cdot \epsilon, 0.05 \cdot \sqrt{3}/2 \cdot \sOR \cdot 1/\sqrt{k} \cdot \epsilon\right) \quad\mbox{($\because C' \geq 1$)} \nonumber
\end{align}
Hence proving the required bound for $\eone$.

Next we prove the $\etwoone$ case. Choose a $C$ that satisfies all previous bounds also obeys $(\sqrt{C}/4)\cdot 0.01\cdot(\sqrt{3}/8) \geq 1$. Note that, since the bound on $\etwoone$ in Lemma~\ref{lem_HKZ_8} is the same as for $\eone$, we can start with the analogue of equation~\eqref{eq:eone_ineq_with_N}:
\begin{align*}
\etwoone &\leq \frac{2\epsilon (\sPtwoone^2\sOR)}{t\sqrt{Ck(1+n_0(\varepsilon)\sPtwoone^2)}}\sqrt{\frac{\ln (3/\eta)}{\ln (1/\eta)}}
\end{align*}
We now drop the additive $n_0(\varepsilon)\sPtwoone^2$ in the denominator, again replace $\sqrt{\ln (3/\eta)/\ln(1/\eta)}$ by $2$ since it is at most $\sqrt{\ln 3}$, and drop the multiplicative factor $t$ from the denominator.
\begin{align*}
\etwoone &\leq \frac{1}{\sqrt{C}}\cdot4\cdot \sPtwoone^2\cdot\sOR\cdot(1/\sqrt{k})\cdot\epsilon \\
&= \frac{1}{\sqrt{C}}\cdot 4\cdot \left[\sPtwoone^2\right]\cdot\left[\sOR/\sqrt{k}\right]\cdot\epsilon \\
&\leq \frac{1}{\sqrt{C}}\min\left(4\cdot\sPtwoone^2\cdot\epsilon, 4\cdot\sOR\cdot \sPtwoone^2 \cdot 1/\sqrt{k}\cdot\epsilon\right) \quad\mbox{($\because \left[\sOR/\sqrt{k}\right] \leq 1$)}\\
&\leq \frac{1}{C'}\min\left(0.05\cdot 1/8 \cdot \sPtwoone^2\cdot\epsilon, 0.01\cdot\sqrt{3}/8 \cdot \sOR\cdot\sPtwoone^2\cdot 1/\sqrt{k}\cdot\epsilon\right) \\
&\quad\mbox{(for $C' = (\sqrt{C}/4)\cdot 0.01\cdot(\sqrt{3}/8)$)} \\
&\leq \min\left(0.05\cdot 1/8 \cdot \sPtwoone^2\cdot\epsilon, 0.01\cdot\sqrt{3}/8 \cdot \sOR\cdot\sPtwoone^2\cdot 1/\sqrt{k}\cdot\epsilon\right) \quad\mbox{(since $C' \geq 1$)}
\end{align*}
hence proving the bound on $\etwoone$.

Finally for $\sum_x \ethreexone$, assume $C$ such that $\frac{2\cdot 0.39\cdot(3\sqrt{3}/8)\sqrt{C}}{16 + \sqrt{C}}\geq 1$ in addition to previous requirements on $C$. we first restate the bound from Lemma~\ref{lem_HKZ_8}:
\begin{align*}
\sum_{x} \ethreexone &\leq \min_{j} \left( \sqrt{j/N}\left(\sqrt{\ln 3/\eta} + 1\right) + 2\epsilon(j)\right) + \sqrt{1/N}\left(\sqrt{\ln 3/\eta} + 1\right)  \\
&\leq \sqrt{n_0(\varepsilon)/N}\left(\sqrt{\ln 3/\eta} + 1\right) + 2\epsilon(n_0(\varepsilon)) + \sqrt{1/N}\left(\sqrt{\ln 3/\eta} + 1\right)  \\
&\leq \sqrt{1/N}\left(\sqrt{\ln 3/\eta} + 1\right)\left(n_0(\varepsilon) + 1\right) + 2\varepsilon \quad\mbox{(since $\epsilon(n_0(\varepsilon)) \leq \varepsilon$)} \\
\end{align*}
The first two terms are exactly as before, so we perform the same steps as in equations~\eqref{eq:samplebound_step1}-\eqref{eq:samplebound_step2} except we do not drop $t\sqrt{n_0(\varepsilon)}$, to get:
\begin{align*}
\sum_{x} \ethreexone &\leq \frac{4 \sOR\sPtwoone^2 \epsilon}{\sqrt{Ckn_0(\varepsilon)}\sPtwoone}\left(n_0(\varepsilon) + 1\right) + 2\varepsilon \\
&\leq \frac{4 \sOR\sPtwoone \epsilon}{t\sqrt{Ckn_0(\varepsilon)}}\cdot (2\cdot n_0(\varepsilon))+ 2\sOR\sPtwoone\epsilon/4t\sqrt{k} \\
&\quad\mbox{(since $1 + n_0(\varepsilon) \leq 2\cdot n_0(\varepsilon)$, and plugging in $\varepsilon$)}\\
&\leq \sOR\cdot\sPtwoone\cdot t\sqrt{k} \cdot\epsilon\cdot\left(8/\sqrt{C}+ 1/2\right)\\
&\leq \frac{1}{C'}0.39\cdot(3\sqrt{3}/8)\cdot\sOR\cdot\sPtwoone\cdot t\sqrt{k}\cdot\epsilon \quad\mbox{(for $C' = \frac{2\cdot 0.39\cdot(3\sqrt{3}/8)\sqrt{C}}{16 + \sqrt{C}}$)}\\
&\leq 0.39\cdot(3\sqrt{3}/8)\cdot\sOR\cdot\sPtwoone\cdot t\sqrt{k}\cdot\epsilon \quad\mbox{(since $C' > 1$ by assumption)}\\
\end{align*}
Hence proving the required bound for $\sum_x \ethreexone$.

\underline{Proof of second step}: Substituting from equation~\eqref{eq:epsmins} into $\delone$ in Lemma~\ref{lem_HKZ_10}:
\begin{align*}
\delone &\leq \frac{2}{\sqrt{3}}\frac{\sqrt{k}}{\sOR }\cdot \eone \\
&\leq \frac{2}{\sqrt{3}}\frac{\sqrt{k}}{\sOR }\min\left(.05\cdot\frac{3}{8}\sPtwoone \epsilon, .05\cdot\frac{\sqrt{3}}{2}\sOR \frac{1}{\sqrt{k}}\epsilon\right) \\
&= .05\epsilon\cdot\min\left(\frac{\sqrt{3}}{4}\frac{\sqrt{k}}{\sOR } \sPtwoone ,1\right) \\
&\leq .05\epsilon
\end{align*}
Substituting from equation~\eqref{eq:epsmins} into $\delinf$ in Lemma~\ref{lem_HKZ_10}:
\begin{align*}
\delinf &\leq  4\left(\frac{\etwoone}{\sPtwoone ^2} + \frac{\eone}{3\sPtwoone }\right) \\
&\leq  \frac{4}{\sPtwoone ^2}\min \biggl(.05\cdot(1/8)\cdot \sPtwoone ^2\cdot(\epsilon/5), .01\cdot(\sqrt{3}/8)\cdot\sOR \cdot \sigma_k (\Ptwoone)^2\cdot(1/(t\sqrt{k}))\cdot \epsilon \biggr) \\
&\quad+ \frac{4}{3\sPtwoone }\min \biggl(.05\cdot(3/8)\cdot\sPtwoone \cdot\epsilon, .05\cdot(\sqrt{3}/2)\cdot\sOR \cdot(1/\sqrt{k})\cdot \epsilon\biggr) \\
&\leq  \min \biggl(.05\epsilon, .04\cdot(\sqrt{3}/8)\cdot\sOR \cdot(1/(t\sqrt{k}))\cdot \epsilon \biggr) \\
&\quad+ \min \biggl(.05\cdot(1/2)\cdot\epsilon, .05\cdot(2/\sqrt{3})\cdot\frac{\sOR }{\sPtwoone }\cdot(1/\sqrt{k})\cdot \epsilon\biggr) \\
&\leq .05\epsilon(.01 + .5) \\
&\leq .05\epsilon
\end{align*}
Substituting from equation~\eqref{eq:epsmins} into $\Delta$ in Lemma~\ref{lem_HKZ_10}:
\begin{align*}
&\Delta \leq \frac{8}{\sqrt{3}}\cdot\frac{\sqrt{k}}{\sOR}\cdot\left(\frac{\etwoone}{\sPtwoone^2} + \frac{\Sigma_x \ethreexone}{3\sPtwoone}\right) \\
\leq &\frac{8\sqrt{k}}{\sqrt{3}\sOR}\cdot\Biggl(\frac{1}{\sPtwoone^2}\min \biggl(.05\cdot(1/8)\cdot \sPtwoone ^2\cdot(\epsilon/5), .01\cdot(\sqrt{3}/8)\cdot\sOR \cdot \sigma_k (\Ptwoone)^2\cdot\frac{\epsilon}{t\sqrt{k}}\biggr) \\ &\quad+ \frac{1}{3\sPtwoone}0.39 \cdot (3\sqrt{3}/8)\cdot\sOR \cdot\sPtwoone \cdot(1/(t\sqrt{k}))\cdot\epsilon\Biggr)\\ &= \Biggl(\min \biggl(.05\cdot(\epsilon/5)\frac{\sqrt{k}}{\sqrt{3}\sOR}, .01\cdot\frac{\epsilon}{t}\biggr) + 0.39 \cdot\frac{\epsilon}{t}\Biggr)\\
&\leq  .01\cdot\frac{\epsilon}{t} +  0.39 \cdot\frac{\epsilon}{t} \\
&\leq 0.4\epsilon/t
\end{align*}

\underline{Proof of third step}: By Lemma~\ref{lem_HKZ_12},
\begin{align*}
\sum_{x_{1:t}}\left|\Pr[x_{1:t}] - \widehat{\Pr}[x_{1:t}]\right| &\leq  (1+\delinf)(1+\delone)(1+\Delta)^t - 1 \\
&\leq (1+.05\epsilon)(1+.05\epsilon)(1+0.4\epsilon/t)^t - 1 \quad \mbox{(by equations~\eqref{eq:delbounds})} \\
&\leq (1+.05\epsilon)(1+.05\epsilon)(1+0.8\epsilon) - 1 \quad \mbox{(by equation~\eqref{eq:a_by_t_ineq}, since $0.4\epsilon < 1/2$)} \\
&= 1 + .05\epsilon + .05\epsilon + .05^2\epsilon + 0.8\epsilon + .04\epsilon^2 + .04\epsilon^2 + (.05)^2 \cdot .08\epsilon^3 - 1\\
&= .0002\epsilon^3 + .0825\epsilon^2 + 0.9\epsilon \\
&\leq (.0002 + .0825 + 0.9)\epsilon \quad\mbox{(since $\epsilon < 1$ by assumption)}\\
&= 0.9827\epsilon \\
&< \epsilon
\end{align*}
This completes the proof of Theorem~\ref{thm_HKZ_6}.
\end{proof}
\subsection{Proof of Theorem~\ref{thm_HKZ_6} for Continuous Observations}
\label{RRHMM-sec-proofs-kde} For continuous observations, we use Kernel Density Estimation (KDE)~\cite{silverman:1986} to model the observation probability density function (PDF). We use a fraction of the training data points as kernel centers, placing one multivariate Gaussian kernel at each point.\footnote{We use a general elliptical covariance matrix, chosen by SVD: that is, we use a spherical covariance after projecting onto the singular vectors and scaling by the square roots of the singular values.}  The KDE estimator of the observation PDF is a convex combination of these kernels; since each kernel integrates to 1, this estimator also integrates to $1$. KDE theory~\cite{silverman:1986} tells us that as the number of kernel centers and the number of samples go to infinity and the kernel bandwidth goes to zero (at appropriate rates), the KDE estimator converges to the observation PDF in $L_1$ norm.  The kernel density estimator is completely determined by the normalized vector of kernel weights; therefore, if we can estimate this vector accurately, our estimate will converge to the observation PDF as well.

Hence our goal is to predict the correct expected value of this normalized kernel vector given all past observations (or more precisely, given the appropriate sequence of past observations, or the appropriate indicative events/features). In the context of Theorem~\ref{thm_HKZ_6}, joint probability estimates for $t$-length observation sequences are effectively the expectation of entries in a $t$-dimensional tensor formed by the outer product of $t$ indicator vectors.  When we move to KDE, we instead estimate the expected outer product of $t$ \emph{stochastic} vectors, namely, the normalized kernel weights at each time step.  As long as the sum of errors in estimating entries of this table goes to zero for any fixed $t$ as the number of samples increases, our estimated observation PDFs will have bounded error.

The only differences in the proof are as follows. In Lemma~\ref{lem_HKZ_8}, we observe $\qvec_i$ to be stochastic vectors instead of indicator vectors; their expectation is still the true value of the quantity we are trying to predict. $\pvec_i$ are also stochastic vectors in that proof. In the proof of Proposition~\ref{prop_HKZ_19}, $p_k$ is an arbitrary stochastic vector. Also, $\qvec_i\tps\qvec_i \leq \pnorm{\qvec_i}{1}  = 1$ now instead of being always equal to 1, and the same holds for $\pvec_i\tps\pvec_i$. Also $\pnorm{\phat_i - \pvec_i}{2} \leq \pnorm{\phat_i - \pvec_i}{1} = 1$ (by triangle inequality). Besides these things, the above proof goes through as it is.

Note that in the continuous observation case, there are continuously many observable operators $W_x$ that can be computed. We compute one base operator for each kernel center, and use convex combinations of these base operators to compute observable operators as needed.
\section{Appendix II: An Example of Learning with Ambiguous Observations}
\label{APPENDIX-sec-ambig} When stacking observations, the modified, larger $\overline{\Ptwoone} \in \Rbbmat{\overline{n}}{\overline{n}}$ still has rank at most $k$ since it can be written in the form $\overline{\Ptwoone} = G T H$ for some matrices $G,H\tps \in \Rbbmat{\overline{n}}{m}$. For example, if $n=2$ for an HMM with ambiguous observations, and we believe stacking $2$ observations per timestep will yield a sufficiently informative observation, the new observation space will consist of all $\overline{n}=n^2=4$ possible tuples of single observations and $\Ptwoone \in \Rbbmat{n^2}{n^2}$, with each observation $i$ corresponding to a tuple $<i_1,i_2>$ of the original observations. Specifically,
\begin{align*}
\overline{\Ptwoone}(j,i) &= \Pr(x_4 = j_2, x_3 = j_1, x_2 = i_2, x_1 = i_1) \\
&= \sum_{a,b,c,d} \Pr(x_4 = j_2, x_3 = j_1, x_2 = i_2, x_1 = i_1, h_4 = d, h_3 = c, h_2=b, h_1 = a) \\
&= \sum_{a,b,c,d} O_{j_2 d} T_{dc} O_{j_1 c} T_{cb} O_{i_2 b} T_{ba} O_{i_1 a} \pi_a\\
&= \sum_{b,c} \overline{O}_{j,c} T_{cb} [\diag(\pi)\overline{O}\tps]_{b,i} \text{\ \ \ \ \ where $\overline{O}_{j,c} = \sum_{d} O_{j_2 d}T_{dc} O_{j_1 c}$} \\
\Rightarrow \overline{\Ptwoone} &= \overline{O}T\diag(\pi)\overline{O}\tps
\end{align*}
Similarly, we can show that $\overline{\Pthreexone} = G T H\tps$ for some matrices $G,H\tps \in \Rbbmat{\overline{n}}{m}$. The exact formulae will differ for different choices of past and future observable statistics.
\section{Appendix III: Synthetic Example RR-HMM Parameters}
\label{APPENDIX-sec-synth}
\paragraph{Example 1}
\begin{align}
T = \left [ \begin{array}{ccc}
     0.3894    &0.2371    &0.3735\\
     0.2371    &0.4985    &0.2644\\
     0.3735    &0.2644    &0.3621
\end{array} \right] \ \ \ \
O = \left [\begin{array}{ccc}
     0.6000   & 0.2000    &0.2000\\
     0.2000    &0.6000    &0.2000\\
     0.2000    &0.2000    &0.6000
  \end{array}  \right]
\end{align}

\paragraph{Example 2}
\begin{align*}
T = \left [ \begin{array}{ccc}
     0.6736  & 0.0051  & 0.1639\\
     0.0330  & 0.8203  & 0.2577\\
     0.2935  & 0.1746  & 0.5784
\end{array} \right] \ \ \ \
O = \left [\begin{array}{ccc}
1& 0 & .5\\
0  & 1 & .5
  \end{array}  \right]
\end{align*}

\paragraph{Example 3}
\begin{align*}
T = \left [ \begin{array}{cccc}
     0.7829    &0.1036    &0.0399    &0.0736\\
     0.1036    &0.4237    &0.4262    &0.0465\\
     0.0399    &0.4262    &0.4380    &0.0959\\
     0.0736    &0.0465    &0.0959    &0.7840\end{array} \right] \ \ \ \
O = \left [\begin{array}{cccc}
1& 0 & 1 & 0\\
0  & 1 & 0 & 1
  \end{array}  \right]
\end{align*}
\end{document}